\documentclass[journal]{IEEEtran}
\usepackage{cite}
\usepackage{multicol}
\usepackage[colorlinks]{hyperref}
\usepackage{bm}
\usepackage{rotating}
\usepackage{url}
\usepackage{graphicx} 
\usepackage{mathptmx} 
\usepackage[mathscr]{euscript}  
\usepackage{amsmath} 
\usepackage{amssymb}  
\usepackage{multirow}
\usepackage{siunitx}
\usepackage{xcolor}
\usepackage{adjustbox}
\usepackage{array}
\usepackage{booktabs}
\usepackage[ruled,vlined]{algorithm2e}  

\usepackage[inline]{enumitem}
\usepackage[hang,flushmargin]{footmisc}
\usepackage{microtype}
\usepackage{threeparttable}
\usepackage[font=footnotesize,labelfont=bf]{caption}
\usepackage{pifont}
\usepackage{mathtools}

\usepackage{verbatim}
\usepackage{xcolor}

\usepackage{tikz}
\usetikzlibrary{arrows, shapes,positioning, shadows,trees}

\tikzset{
  basic/.style  = {draw, text width=4cm,  rectangle},
  root/.style   = {basic, rounded corners=2pt, thin, align=center},
  level 2/.style = {basic, thin, rounded corners=2pt, align=center,text width=3.7cm},
  level 3/.style = {basic, thin, rounded corners=2pt, align=left, text width=2.8cm}
}

\pdfoutput=1

\newcommand{\secref}[1]{Section~\ref{#1}}
\renewcommand{\eqref}[1]{Equation~(\ref{#1})}
\newcommand{\figref}[1]{Figure~\ref{#1}}

\newcolumntype{P}[1]{>{\centering\arraybackslash}p{#1}}

\newcommand{\change}[1]{\textcolor{black}{#1}}

\begin{document}

\title{Fairness and Bias in Robot Learning}
\author{Laura Londoño$^{*,1}$,
        Juana Valeria Hurtado$^{*,1}$,
        Nora Hertz$^2$,
        Philipp Kellmeyer$^{3,4}$,\\
        Silja Voeneky$^2$,
        and~Abhinav Valada$^1$
\thanks{$^*$These authors contributed equally.}%
\thanks{$^1$Department of Computer Science, University of Freiburg, Germany}
\thanks{$^2$Dep. International Law and Ethics of Law, University of Freiburg, Germany}%
\thanks{$^3$School of Business Informatics and Math., University of Mannheim, Germany.}%
\thanks{$^4$Department for Neurosurgery, University Medical Center Freiburg, Germany}}

%



\maketitle

\begin{abstract}
Machine learning has significantly enhanced the abilities of robots, enabling them to perform a wide range of tasks in human environments and adapt to our uncertain real world. Recent works in various machine learning \change{domains} have highlighted the importance of accounting for fairness to ensure that these algorithms do not reproduce human biases and consequently lead to discriminatory outcomes. With robot learning systems increasingly performing more and more tasks in our everyday lives, it is crucial to understand the influence of such biases to prevent unintended behavior toward certain groups of people. In this work, we present the first survey on fairness in robot learning from an interdisciplinary perspective spanning technical, ethical, and legal challenges. We propose a taxonomy for sources of bias and the resulting types of discrimination due to them. Using examples from different robot learning domains, we examine scenarios of unfair outcomes and strategies to mitigate them. We present early advances in the field by covering different fairness definitions, ethical and legal considerations, and methods for fair robot learning. With this work, we aim \change{to pave} the road for groundbreaking developments in fair robot learning.
\end{abstract}


%
\IEEEpeerreviewmaketitle

\section{Introduction}

\IEEEPARstart{R}{obot} learning has advanced tremendously in the last decade. From learning low-level manipulation skills~\cite{peters2013towards, shankar2020learning, hausman2018learning} to long-horizon mobile manipulation tasks~\cite{blomqvist2020go, honerkamp2021learning}, and autonomous driving~\cite{cattaneo2022lcdnet, sirohi2021efficientlps, valada2016convoluted}, machine learning has accelerated the advancement in the entire spectrum of robotic domains. Much of this success has been fueled by data-driven learning algorithms, massive curated datasets, and the doubling of computational capacity each year. We also \change{witness} more and more learned robotic systems performing tasks in human-centered environments alongside humans. Notable areas include robots in collaborative manufacturing~\cite{asfour2018armar}, agriculture~\cite{kohanbash2012base}, logistics~\cite{cattaneo2020cmrnet}, and search and rescue~\cite{mittal2019vision}. Along with these technical advances, studying the ethical as well as legal implications and accounting for fairness in machine learning algorithms is growing in concern. Several recent studies in different areas such as natural language processing~\cite{caliskan2017semantics, jacobs2020meaning, chang2019bias}, facial recognition systems~\cite{gebru2019oxford, raji2020saving, denton2019detecting}, and risk assessment systems~\cite{dressel2018accuracy, tolan2019machine} have demonstrated the vulnerability of these algorithms to biases which leads them to exhibit discriminatory behaviors. Nevertheless, fairness remains little explored in the context of robot learning. \change{Machine learning primarily deals with data-driven algorithms such as image recognition and recommendation systems, where bias can manifest in data and algorithms. In contrast, robot learning also focuses on the adaptation of the robot's behavior and their interaction with the environment and with humans, posing additional ethical concerns related to ensuring socially acceptable behavior and human safety. Accordingly, a robot can be defined as an embodied system capable of carrying out a complex series of actions autonomously.} Robots as social agents can potentially replicate and even amplify human biases such as favoring particular groups of people or disfavoring interaction with specific users, frequently the less represented in the learning process. Fairness in robot learning is a critical area of research that will ensure safety and comfort around humans \change{and} enable ethical decision making. Moreover, preventing and diminishing bias is essential for robot acceptance in human environments and under an appropriate legal framework. 

\begin{figure}
    \footnotesize
    \centering
    \includegraphics[width=\linewidth]{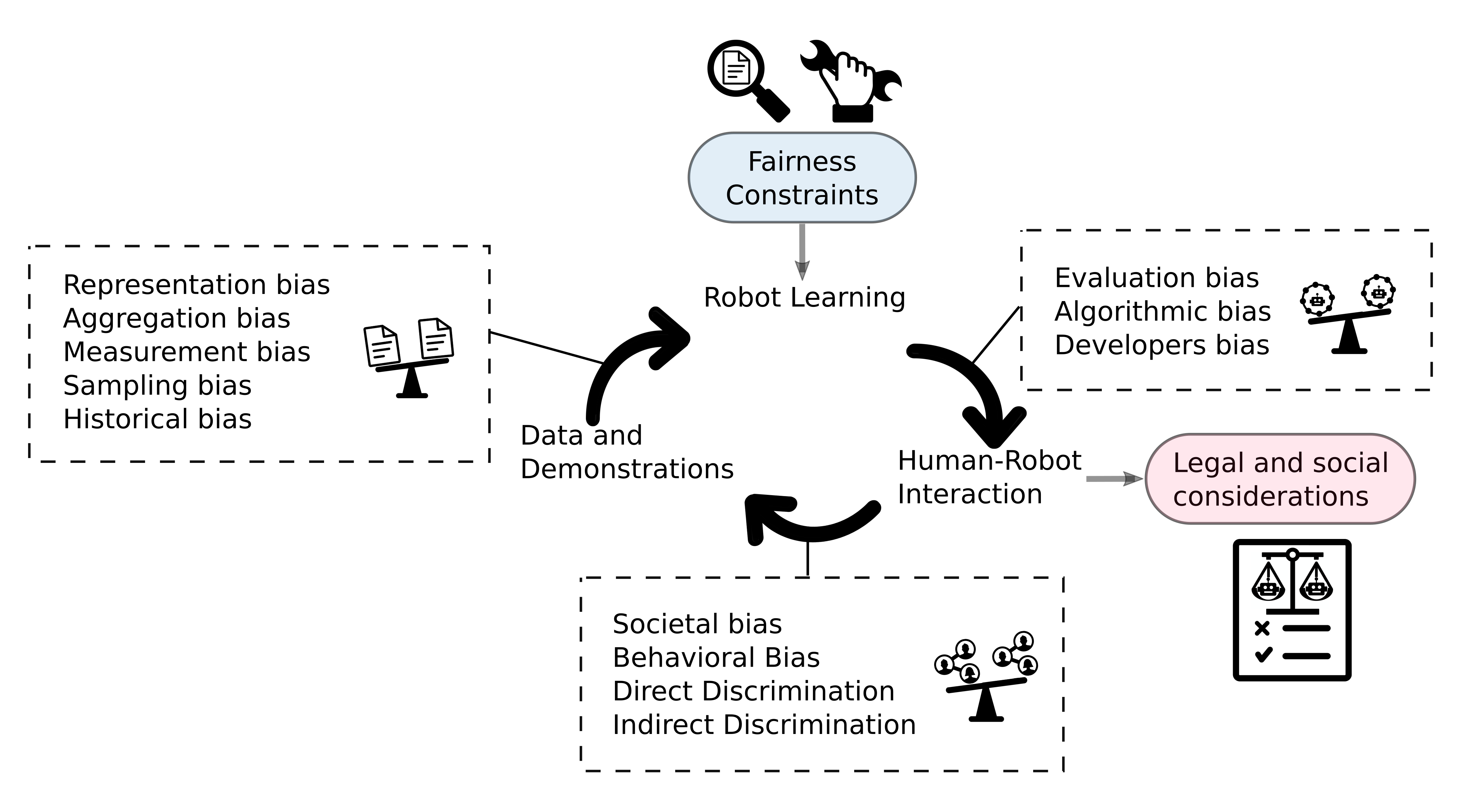}
    \caption{The typical robot learning paradigm consists of different stages. Each of these stages can incorporate bias and result in discriminatory behavior. Accounting for fairness in robot learning is essential for robots to operate in human environments.}
    \label{fig:intropic}
\end{figure}

Different technical factors \change{influence a robot's behavior during the development and deployment stages} as shown in \figref{fig:intropic}. Unforeseen outcomes such as \change{negative} bias and discrimination require detailed investigation from a multidisciplinary perspective. Learning algorithms can leverage information to obtain an optimized model to perform the required task. At the same time, the learned model can encode bias through different sources such as societal, historical, measurement, and representation bias. Understanding the sources and types of bias in robot learning is crucial for creating strategies towards fairness-aware learning. Additionally, it is essential to determine the requirements to comply with the legal regulations.

The ethical research carried out thus far in robotics studies the consequences of deploying robots in social environments \change{and} the consequences of interacting with humans. The ethical dimension is part of the practical ethics field and its analysis involves social sciences, philosophy, and psychology. Based on the idea that solutions to bias and discrimination in robotics can be \change{not only} technical but also ethical, researchers seek to reflect on how to build fair interactions between humans and robots, the economic and social implications of deploying robots in social settings, and more importantly, what it means for a machine to behave ethically in crowded or personal environments. 

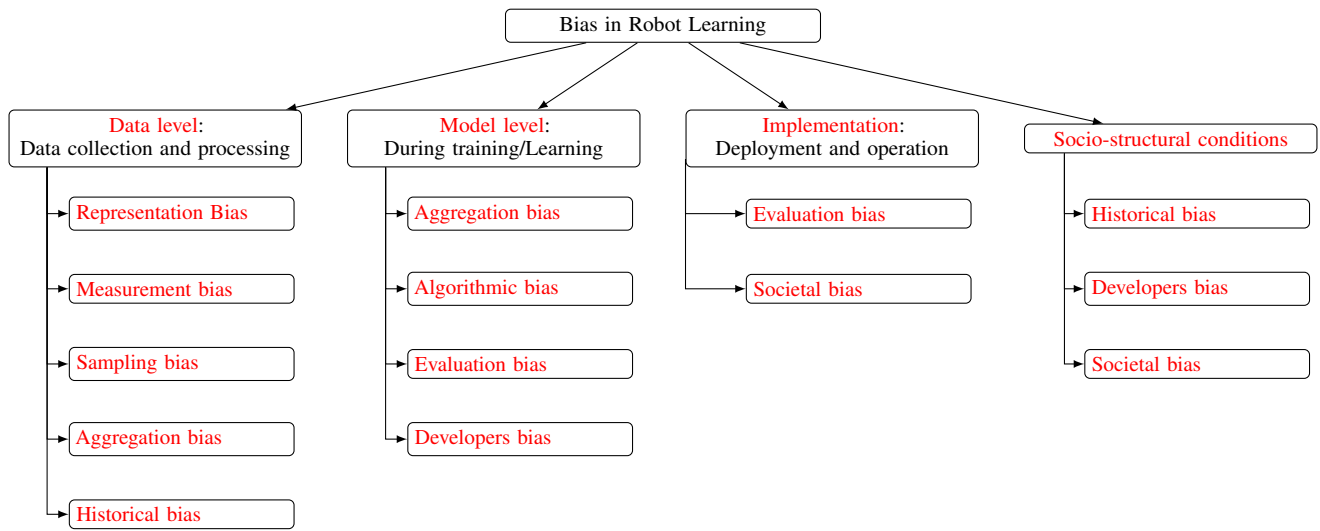
\begin{figure*}
\centering
\footnotesize
\begin{tikzpicture}[
  level 1/.style={sibling distance=45mm},
  edge from parent/.style={->,draw},
  >=latex]

\node[root] {Bias in Robot Learning}
  child {node[level 2] (c1) {\hyperref[sec:introdatalevel]{Data level}: \\ Data collection and processing}}
  child {node[level 2] (c2) {\hyperref[sec:intromodellevel]{Model level}: \\ During training/Learning}}
  child {node[level 2] (c3) {\hyperref[sec:introimplementationlevel]{Implementation}: \\ Deployment and operation}}
  child {node[level 2] (c4) {\hyperref[sec:introsociallevel]{Socio-structural conditions}}};

\begin{scope}[every node/.style={level 3}]
\node [below of = c1, xshift=10pt] (c11) {\hyperref[sec:representation]{Representation Bias}};
\node [below of = c11] (c12) {\hyperref[sec:measurement]{Measurement bias}};
\node [below of = c12] (c13) {\hyperref[sec:sampling]{Sampling bias}};
\node [below of = c13] (c14) {\hyperref[sec:aggregation]{Aggregation bias}};
\node [below of = c14] (c15) {\hyperref[sec:historical]{Historical bias}};

\node [below of = c2, xshift=10pt] (c21) {\hyperref[sec:aggregation]{Aggregation bias}};
\node [below of = c21] (c22) {\hyperref[sec:algorithmic]{Algorithmic bias}};
\node [below of = c22] (c23) {\hyperref[sec:evaluation]{Evaluation bias}};
\node [below of = c23] (c24) {\hyperref[sec:developer]{Developers bias}};

\node [below of = c3, xshift=10pt] (c31) {\hyperref[sec:evaluation]{Evaluation bias}};

\node [below of = c31] (c32) {\hyperref[sec:societal]{Societal bias}};

\node [below of = c4, xshift=10pt] (c41) {\hyperref[sec:historical]{Historical bias}};
\node [below of = c41] (c42) {\hyperref[sec:developer]{Developers bias}};
\node [below of = c42] (c43) {\hyperref[sec:societal]{Societal bias}};
\end{scope}

\foreach \value in {1,...,5}
  \draw[->] (c1.195) |- (c1\value.west);

\foreach \value in {1,...,4}
   \draw[->] (c2.195) |- (c2\value.west);

\foreach \value in {1,2}
  \draw[->] (c3.188) |- (c3\value.west);

\foreach \value in {1,...,3}
  \draw[->] (c4.188) |- (c4\value.west);
  
\end{tikzpicture}
\caption{\change{Categorization of bias in robot learning and the relationship to different sources of bias. Please click on each topic to jump to the corresponding section.}}
\label{fig:tree}
\end{figure*}

In this work, we present an interdisciplinary overview of bias and fairness in robot learning. Our main goal is to provide a comprehensive review that explores technical, ethical, and legal considerations on the topic. We discuss early advances made in this area based on these three fields. We present ethical guidelines, and social and legal considerations, and propose a taxonomy of the types and sources of bias and discrimination in robot learning. Furthermore, we present different fairness definitions, metrics, and methods for fairness-aware robot learning. \change{Based on the three typical stages in robot learning, we categorize bias into four levels: data level (data collection and processing), model level (during training and learning), implementation level (deployment and operation), and socio-structural conditions. We associate sources of bias to each category for clarity as shown in Figure~\ref{fig:tree}. Additionally, we describe techniques for bias detection and mitigation for each of these levels.} As a part of practical ethics, ethics in robotics or roboethics is a multidisciplinary field that aims to find practical ethical solutions to bias and discrimination in robotics. This survey also presents the advances made so far in this field and its applications. Finally, we also present initial practices for fair robot learning. By compiling the ethical, legal, and technical approaches from this recent research topic, we aim to present an initial starting point for interdisciplinary advances toward fairness-aware robot learning.

\change{This survey is structured as follows: First, we outline the paper selection methodology in Section~\ref{sec:methodology}. Subsequently, we discuss the differences between fairness in machine learning and robot learning in Section~\ref{sec:farinessMLvsRL}. In Section~\ref{sec:biasAndDiscri}, we provide a detailed review of the types and sources of bias and discrimination in robot learning. We then discuss the challenges of fairness in robot learning in terms of strategies for detecting and mitigating bias. We also survey fairness metrics and frameworks for evaluation in Section~\ref{sec:tech}. In Section~\ref{sec:legalandfair}, we present relevant aspects of fairness in robot learning from legal and social perspectives. Section~\ref{sec:practices} provides additional practices to consider for fair robot learning. We finalize our survey in Section~\ref{sec:conclusion} presenting concluding remarks.}

\section{Survey Methodology}
\label{sec:methodology}

\change{In this section, we outline the methodology that we used to select papers for our survey. \change{We aim} to cover the intersection of fairness, bias, ethics, and legal considerations in robotics, robot learning, machine learning, and human-robot interaction. We applied the following criteria:}

\begin{itemize}
    \item \change{Bias sources and discrimination criteria: We conducted a literature search on bias and discrimination in machine learning, combining it with relevant literature on bias in social robotics. We then analyzed these sources of bias and the resulting discrimination scenarios from the perspective of robot learning. The keywords used in this search were: bias in machine learning, bias in robotics, bias in robot learning, discrimination in machine learning, discrimination in robotics, and discrimination in robot learning. We studied papers that were published within the last 15 years. We used databases such as Google Scholar, IEEE, ACM, Scopus, Taylor and Francis Online.}
    \item \change{Human-Centric Relevance: We included papers \change{examining fairness and bias within human-robot interaction, focusing on protecting} the human being (required by ethical and/or legal standards). However, we excluded papers solely focused on physical robot attributes such as color or shape. \change{We emphasized} understanding and mitigating bias and unfairness in interactions between robots and humans or the social impact of robots' behavior, rather than the physical characteristics of robots themselves.}
    \item \change{Machine learning applied to Robot learning: Recognizing that bias and fairness have been extensively explored in the general field of machine learning, we included relevant papers that had the potential to be applied to robot learning. These papers were included if they offered insights, techniques, or methodologies that could be adapted or integrated into the development and training of robots to promote fairness and reduce bias in human-robot interaction.}
    \item \change{Ethical considerations in robot learning: We explored relevant ethical guidelines for robotics, focusing on human-robot interactions. When selecting relevant papers, we conducted keyword searches using terms such as roboethics, responsible robot systems, ethics of robotics, philosophy of AI, and philosophy in robotics. We also focused on searching for papers that explain the role of philosophy in the development and behavior of robots. We studied papers published within the last 15 years.}
    \item \change{Legal analysis: We used databases such as SSRN, Taylor \& Francis Online, and HeineOnline and searched for the following keywords: discrimination, fairness, robotics, and regulation. We excluded articles on product safety. We did not restrict our search to a specific period, but most papers were published in the last 20 years. In total, we selected 35 articles for our analysis.}
\end{itemize}

\section{\change{Fairness in Machine Learning vs.\\Robot Learning}}
\label{sec:farinessMLvsRL}

\change{While Machine Learning (ML) and robot learning share similarities, they address different tasks with unique characteristics. We distinguish robot learning models from other ML models by their operational context and summarize these differences in Table~\ref{tableML_RL}. Traditional ML primarily operates on non-embodied systems, and Human-Computer Interaction (HCI) is typically characterized as static. In contrast, robot learning focuses on embodied systems, that often exhibit autonomous execution of actions and the capacity for dynamic interactions with their environment and with humans. Specifically, ML models are primarily data-driven algorithms that learn patterns and make predictions from the input data. They are often used for image recognition, natural language processing, recommendation systems, etc. In Human-Computer Interaction (HCI), ML enhances user experience by personalizing content and enhancing system performance.}

\change{In contrast, robot learning models enable robots to adapt to their environments and interact with humans. Besides data analysis and prediction, robot learning focuses on the ability of robots to make decisions and take actions in the real world. In Human-Robot Interaction~(HRI), robot learning plays a critical role in enabling robots to be more socially aware and capable of collaborative tasks with humans. Robot learning models can also be developed for physical interaction with humans. They learn not only from data but also from human guidance and feedback. In HRI, robot learning is essential for enabling robots to adapt their behavior based on social cues, human preferences, comfort, and safety considerations during physical interactions. Addressing ethical and social concerns, ML models deal with bias, privacy, and fairness, mainly when deployed in decision making systems. Robot learning models raise additional ethical and safety considerations such as ensuring that robots behave in a socially acceptable manner, respecting personal boundaries, and prioritizing human safety during physical interactions. HRI research focuses on these aspects to create robots that enhance human well-being and safety.}


\begin{table*}
\centering
{\color{black}\begin{tabular}{p{4.2cm}|p{6.5cm}|p{6cm}}
\toprule
\textbf{Aspect} & \textbf{Machine Learning} & \textbf{Robot Learning} \\
\midrule
\textbf{Operational Context }& Non-embodied systems & Embodied systems \\
\textbf{Functionality and Tasks} & Data-driven algorithms for predictions and decisions & Enable robots to learn, adapt, and interact physically \\
\textbf{Interaction Purpose }& Operate in the background with minimal user interaction & Engage in physical interaction and adapt to humans \\
\textbf{Interaction Dynamics }& Operate in a more static and restrictive environment & Option to be autonomously mobile and interact in dynamic environments \\
\textbf{Learning Methods }& Supervised, unsupervised, reinforcement learning & Reinforcement learning, imitation learning, and human-in-the-loop learning \\
\textbf{Fairness and Social Considerations} & Ethical concerns related to bias, fairness, and privacy & Additional concerns regarding social acceptability, personal boundaries, human comfort and safety \\
\bottomrule
\end{tabular}}
\caption{\change{Comparison of key characteristics of machine learning and robot learning models.}}
\label{tableML_RL}
\end{table*}

\section{Types and Sources of Bias and Discrimination}
\label{sec:biasAndDiscri}

From a societal perspective, "bias" can be defined as the tendency to weigh disproportionately in favor or against groups of people, causing prejudicial or discriminatory situations. Similarly, in machine learning and AI, bias refers to when a model systematically and unfairly treats \change{specific} individuals or groups of individuals in favor of others~\cite{friedman1996bias}. \change{As a result, a series of discriminatory situations have arisen due to unintended and intended bias that has affected numerous people. Discrimination is the final act or result of bias. Contrary to “bias”, discrimination is a well-known legal concept and most countries know the anti-discrimination legislation~\cite{kerrigan2022artificial}.} For example, discrimination based on sex/gender~\cite{cirillo2020sex,gebru2019oxford,caliskan2017semantics}, discrimination in algorithms for job recruitment~\cite{krishnakumar2019assessing, wilfred2018ai}, and many other contexts. This issue has led to widespread concern as it perpetuates historical injustices and inequitable social structures. Thus, researchers from multidisciplinary fields have argued that it is essential to identify its causes to prevent \change{discrimination} in machine learning and AI. We can observe a similar trend in the field of robot learning. As social robots and other autonomous systems such as self-driving cars are increasingly being developed, it is crucial to set out for a comprehensive understanding of sources of unwanted consequences, \change{namely bias}~\cite{howard2018hacking, ball2005gender, howard2020robots, bartneck2018robots, hurtado2022feminist}.

In this section, we present a review to recognize potential types and sources of bias in the field and discriminatory situations in robot learning. \change{Importantly, in many of the scenarios described, we do not exclusively find a single bias at work but many of the biases described here are interconnected, both conceptually and in how they may impact actual robot learning. Despite this substantial overlap, we find that a taxonomy of the various biases is nevertheless important as this provides the conceptual basis for designing and justifying specific interventions targeted at mitigating specific biases. Ideally, some of these interventions may mitigate several biases that work in concert at once. In many cases, however, targeting specific biases may also help for overall bias mitigation.}

\subsection{Bias in Robot Learning}

Most robot learning algorithms follow the data-driven paradigm. This allows robots to automatically learn from data using guidance or supervision to optimize the models for specific tasks such as navigation. For instance, supervised learning approaches utilize datasets containing data and the corresponding annotations. This annotated data is then used to learn a model that generates the desired output. Other learning approaches, such as reinforcement learning and imitation learning, do not require the labeled input-output data pairs. Instead, these approaches use the experiences of an agent performing actions as the learning guidance. In reinforcement learning, the agent takes actions in an environment, and a reward function punishes or encourages the decisions to maximize the notion of cumulative reward to obtain an optimal model. With inverse reinforcement learning approaches, imitation learning aims to optimize a model to generate actions that imitate an example behavior. In this case, instead of using a reward, the reward function is inferred from a set of expert examples. The idea is to mimic by recovering a cost function that explains the observed behavior~\cite{billing2010formalism, brys2015reinforcement}.

\change{Since these approaches use the data-driven paradigm, the first natural source of unfair results in robot learning is data containing human biases or datasets without fairly represented samples. Learning algorithms can significantly replicate and even amplify unfair situations explicitly or implicitly included in the utilized data, which constitutes a data level bias as described in \secref{sec:introdatalevel}. In addition to data-related bias, learning approaches are susceptible to finding over-associations to features related to a particular characteristic such as race, age, or gender, and the prediction task, which constitutes a model level bias as detailed in \secref{sec:intromodellevel}. Furthermore, the intermediate representations of learned models are challenging to interpret, and the models are often considered black boxes where understanding the inside representations is not a priority. Therefore, learned models have challenges regarding explainability, leading to challenges in determining the reasoning behind biased outputs. We categorize these challenges that make robot learning prone to yielding biased behavior after deployment as implementation-level bias, described in \secref{sec:introimplementationlevel}. Moreover, it is important to acknowledge that, as humans naturally shape their social perceptions in diverse ways, socio-structural conditions can permeate the robot learning process, introducing an additional dimension of bias as described in \secref{sec:introsociallevel}. \figref{fig:tree} illustrates this categorization of bias in robot learning and its relationship to the different sources of bias.}\looseness=-1

\subsection{Data Level Bias}
\label{sec:introdatalevel}

\begin{figure}
    \footnotesize
    \centering
    \includegraphics[width=0.47\textwidth]{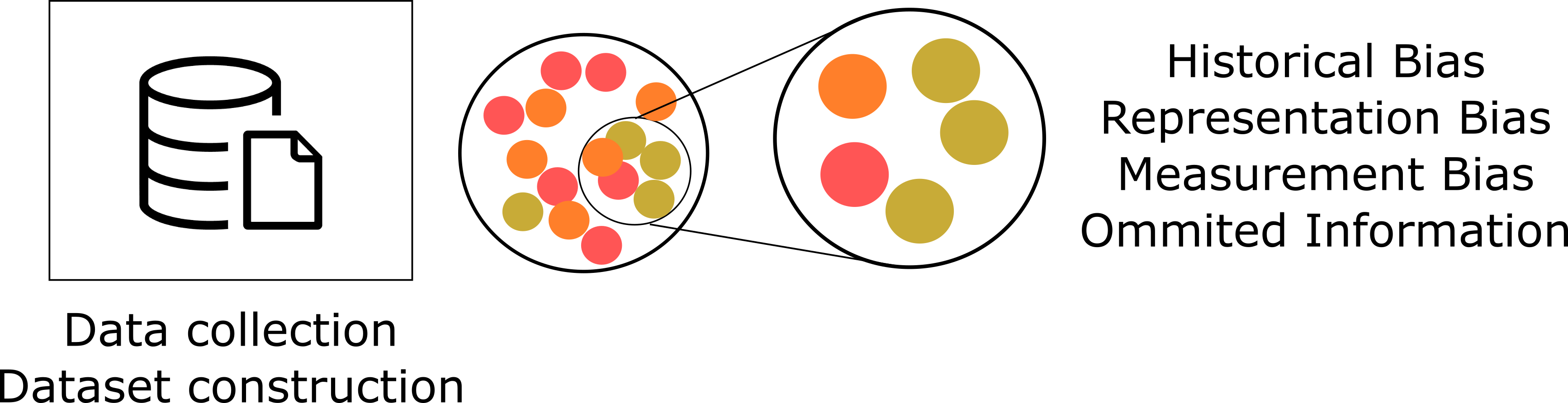}
    \caption{The data used to learn the model can contain social bias. This data level bias in robot learning includes datasets with historical bias, i.e.,  data and information that do not accurately represent relevant details of the real world or contain measurement errors, among others.}
    \label{fig:datalevel}
\end{figure}

Most learning approaches require a large amount of data so that the model learns to yield the desired output. In robotics, the data is used to learn models for navigation, decision making, manipulation, and other tasks. This data is collected as samples from the real world, simulations, or recordings during experiments in controlled environments. In the case of data gathered from the real world, the process itself can reproduce biases, reflecting human biased behaviors. \change{The possible situations to simulate are limited in the case of data collected from simulation and control experiments.} Therefore, only limited segments of the real-world scenarios are included in the learning process, and deploying a robot in new environments is challenging due to the mismatch in data distribution. Additionally, when using human behavior and interaction data, it is also essential to collect a representative sample in a dataset that accurately depicts the general population. Training a model with data that misrepresent the population leads to optimization for the more represented groups to increase the general model performance, leading to poor prediction accuracy for the less represented group. The data used for robot learning is generally sensitive to historical human biases and incomplete or unrepresentative training data, as illustrated in \figref{fig:datalevel}.

\change{Some works} include fairness-sensitive features in robot learning algorithms. For example, \cite{patompak2020learning} uses a gender-specific behavior for navigation. Subsequently, \cite{hurtado2021learning} analyzes the challenges of using these explicit features regarding social implications. Most robot learning algorithms do not explicitly use the so-called "protected" attributes to train the model. For example, using “age” as an attribute to learn navigation models is not common. On the other hand, it is more common that the learned models rely on raw input data. Nevertheless, the absence of explicitly encoded attributes in the input does not protect the output against bias related to these attributes. Unwanted discrimination can result in high relations between features and the task goal. For instance, even if we do not use gender as input to train a navigation model from human demonstrations, the model may learn “notions” of gender from the implicit information in the data.
On top of that, data disparity can occur even if we do not use protected attributes related to an underrepresented group. Since the learned model aims to fit the training data and any training data is prone to contain bias, the models are susceptible to replicating the biases in data. Furthermore, learning models can also amplify societal stereotypes by using associations that might result in accurate but unfair performance. Even though bias is not always designated as negative, it is unfair when there is no valid reason for different outputs. Therefore, in data level bias, the model biased output results from bias in the input data.
\subsection{Model Level Bias}
\label{sec:intromodellevel}

In many robot learning scenarios, detecting bias in the input data is challenging. For instance, if we use human demonstrations of navigation trajectories to learn robot navigation, the resulting model can encode the gender attribute. The unwanted attribute encoding is prone to emerge in the model even if humans can not differentiate a feature such as gender, solely from trajectory data. With the resulting model, the robot might make decisions based on the implicitly learned attributes and produce unfair outcomes. Different factors such as redundant encodings or design choices during the robot learning process can lead to the model containing associations between specific robot actions and protected attributes. Therefore, a model that relies on fairness-sensitive features is susceptible to cause discrimination. As a result, the learned model itself is learned to retain encoded biased representations.

Formally, in model level bias, the learned representation captures a bias-related attribute, which later contributes to the model prediction. This bias also prevents the model from learning richer representations since it incorrectly limits the optimization towards unwanted input-output associations. For example, a robot vision system that learns to identify pedestrians and relies on detecting specific skin colors~\cite{wilson2019predictive}. In such cases, it is ideal that the prediction outcome discrimination is detected and removed from the model perspective. Approaches to mitigate model level bias are divided into in-processing and implementation. These approaches recognize that modeling techniques can lead to biased outputs by dominant features, and distributional effects, or aim to produce accurate and fair models by finding a balance between accuracy and fairness model objectives. Moreover, model level techniques tackle situations in which novel scenarios or situations appear, and it is necessary to adapt the learned model. These approaches tackle this challenging problem by often incorporating one or more fairness metrics into the model optimization function. Therefore, the goal is to maximize both performance and fairness notions.

\change{\subsection{Implementation Bias}
\label{sec:introimplementationlevel}
Implementation bias refers to the potential bias that can occur during the deployment and operation of a robot in real-world settings. While data level and model level biases are primarily concerned with biases present in the training data and the learned models, implementation bias deals with the bias that can arise when these models are put into practical use.
The transition from a controlled environment, where training data is collected, to real-world deployment can introduce a range of challenges and biases. Examples of these are:}

\begin{itemize}
\item \change{Explainability and Interpretability: When a robot exhibits unfair behavior, it can be challenging to precisely determine the reasons behind its actions. Robot learning models, particularly complex ones, often lack transparency, making it difficult to discern the exact decision making process that led to unfair outcomes~\cite{omeiza2021explanations, omeiza2021explanations}.}

\item \change{Changes in the environment: Real-world environments are often more complex and unpredictable than controlled training environments or simulations~\cite{de2023practical}. These discrepancies can lead to biases in system performance as the model may not have been adequately exposed to the diversity of scenarios it encounters during deployment.}

\item \change{User Interaction Bias: How users interact with the robot can also introduce bias. User behaviors, preferences, and cultural differences can affect the robot's response, potentially leading to unfair or unintended outcomes~\cite{hurtado2021learning}.}

\item \change{Domain Adaptation Challenges: The real-world context is dynamic, requiring the model to adapt~\cite{vodisch2023codeps,bevsic2022unsupervised}. The process of adaptation itself can introduce bias if not carefully managed, as the model may make adjustments based on recent data that reinforce existing biases.}

\item \change{Feedback Loops: Biases can be perpetuated through feedback loops~\cite{pagan2023classification, stray2023ai, smith2023bias} . For example, if the system is used to interact with a particular group more frequently, it may become increasingly tailored to that group's needs, unintentionally neglecting the needs of underrepresented groups.}
\end{itemize}

\subsection{Socio-structural conditions}
\label{sec:introsociallevel}

\change{It is well known that humans themselves are the main sources of bias. These biases are often regarded as inherent characteristics of humans. Therefore, they have been analyzed from both cognitive and anthropological perspectives. Generally, those biases resulted in people's judgments and decisions regarding the world and people around them \cite{dror2008cognitive}. These judgments can be of a positive or negative nature and can be shared by large groups of people with similar characteristics. When they are negative, it results in cases of discrimination, segregation, and intolerance towards certain populations. Then, a series of structural inequalities in society emerge. Simultaneously, these biases also affect social institutions, practices, and attitudes~\cite{lopez2021bias}. It is introduced into a robot learning system either via a prejudiced individual, consciously or unconsciously, or via society at large.} 

\subsection{Sources of Bias}

\begin{table*}
\centering
{\color{black}\begin{tabular}{p{2.8cm}|p{11cm}|p{3cm}}
\toprule
\textbf{Type / source of bias} & \textbf{Description} & \textbf{Context} \\
\midrule
\hyperref[sec:representation]{\textbf{Representation bias}}& Occurs when the definition and sampling of a population, such as in geographical areas, leads to non-uniform distribution in variables used for robot autonomy, resulting in biased decisions that can affect certain groups more than others~\cite{mehrabi2019survey}. & Data \\
\hyperref[sec:measurement]{\textbf{Measurement bias}} & Arises when the data collected for training significantly differs from the data obtained during real-world deployment~\cite{mehrabi2019survey} or when defective measures cause skewed data, leading to biases in robot decision making. & Data \\
\hyperref[sec:sampling]{\textbf{Sampling bias}}& Occurs when data selection does not achieve adequate randomization, making some members of the population or specific environmental characteristics less likely to be included in the sample, resulting in an inaccurate distribution of training data that does not resemble the actual environment~\cite{landers2023auditing}. & Data \\
\hyperref[sec:historical]{\textbf{Historical bias}}& Refers to the bias arising from socio-cultural inequalities contingent on history and is represented in historical data used for training algorithms, leading to unintended and unforeseen consequences and perpetuation of historical prejudices~\cite{hurtado2021learning}. & Data / Socio-structural conditions \\
\hyperref[sec:aggregation]{\textbf{Aggregation bias}} & Occurs when erroneous conclusions are made for one group based on observations from other groups~\cite{hagendorff2023speciesist}, such as the misidentification of people of color by pedestrian recognition systems in self-driving cars due to a lack of diverse data. & Data / Model \\
\hyperref[sec:algorithmic]{\textbf{Algorithmic bias}} & Concerns algorithms that reflect systematic and unfair discrimination~\cite{hagendorff2020ethical}, even though algorithms are often treated as neutral and unbiased. & Model \\
\hyperref[sec:evaluation]{\textbf{Evaluation bias}} & Arises when benchmarks are used to evaluate algorithms disproportionately and incorrectly measure performance, leading to biased robot decision making, such as robots maintaining greater distance with women than with men~\cite{hurtado2021learning}. & Model / Implementation \\
\hyperref[sec:developer]{\textbf{Developers bias}} & Arises from the implicit and explicit biases of the developers or creators, influenced by their individual beliefs, actions, and cultural and societal context, affecting how models are adapted~\cite{johnson2020algorithmic}. & Model / Socio-structural conditions \\
\hyperref[sec:societal]{\textbf{Societal bias}} & Occurs when judgments or actions of people influence the behavior of robots and the decisions they make, arising from learning interaction behavior from data~\cite{friedman1996bias}. & Implementation / Socio-structural conditions \\
\bottomrule
\end{tabular}}
\caption{\change{Summary of the source of bias definitions and the relationship with the standard robot learning stages.}}
\label{tableSourceofBias}
\end{table*}

The sources of bias have been primarily studied in fields such as machine learning and AI. Recent works have analyzed data biases~\cite{mehrabi2019survey, suresh2019framework, olteanu2019social} presenting complete lists of the different types of biases and their corresponding definitions. Whereas, in robot learning, there have been no detailed and extensive studies on the sources of biases. The presence of robots in social contexts has started to generate studies about the possible sources of bias and its impacts. Recent studies regarding biases in social robotics highlight potential sources of bias in this field, as well as their consequences~\cite{mieczkowski2019helping, sabanovic2006robots, you2011robot, grinbaum2017ethics, hurtado2022feminist}. These studies demonstrate the critical need to evaluate human-robot interaction as a sociocultural activity. Therefore, recognizing the benefits or prejudices it may generate \change{should be integral to} responsible development. \change{Multiple factors such as data, user behavior, training models, or learning approaches, can be sources of bias.} In the following, we categorize the sources of biases in robot learning and their effects on the functioning of robots as well as the effects on society. \change{Some of those biases can come from similar sources and produce similar effects. Although the sources of bias come from different problems, the results can be similar}.

\subsubsection{Representation Bias}\label{sec:representation} \change{occurs because developers define and sample from a population based on certain general characteristics of specific groups of people, such as their geographical area.} For instance, \cite{brandao2020fair} shows such a simulated scenario where a rescue robot uses geographic population data from the census to search for victims who might need medical assistance after an earthquake. As the spatial distribution of the population within the census data is based on variables such as age, ethnicity, and gender, the non-uniform distribution leads to bias in the variables used for path planning, resulting in biased decisions that can affect \change{specific} populations more than others. In such extreme cases, representation bias leads to inequality, which could put people's lives at risk.

\subsubsection{Measurement Bias}\label{sec:measurement} arises when the data collected for training (testing phase) significantly differs from the characteristics of the social context where the robot is deployed~\cite{suresh2019framework, mehrabi2019survey,van2013designing, ikuta2003safety, sharkey2012granny, hersh2015overcoming}. \change{There is a growing interest in developing healthcare robots for use in healthcare environments such as hospitals. One of those robots may be developed to monitor patients' vital signals such as blood pressure and temperature. However, during the development phase, the robot was primarily tested on healthy individuals}. \change{When the robot is deployed in a real-world healthcare setting, it can encounter a diverse patient population, including people with severe health problems. The robot may provide inaccurate measurements for these diverse patients in this scenario.} 

\subsubsection{Sampling Bias}\label{sec:sampling} arises from how data is selected and does not achieve adequate randomization, making some \change{population members} or \change{specific} environmental characteristics less likely to be included in the sample \cite{johnson2020algorithmic}. As a result, inaccurate \change{training data} distribution does not resemble the \change{actual} environment where robots will be deployed. As each population and social environment has its characteristics, the algorithms in robots should be able to recognize these characteristics.

\subsubsection{Historical Bias}\label{sec:historical} refers to the bias arising from sociocultural inequalities that are contingent on \change{historical events and are reflected in the historical data used to train algorithms. This type of bias originates from preconceived notions or prejudices. Even when data is accurately measured and sampled, learning robotic skills and behavior from this historical real-world data can lead to unintended and unforeseen consequences~\cite{hellstrom2020bias, cowgill2019bias}}. \change{Furthermore, biases from the past can be perpetuated into the future through robots deployed in human environments.} For example, in \cite{Trustrobots} the authors present an example of how historical bias can be transferred through algorithms in robotics. They argue that a package delivery robot trained with \change{spatial data similar to that used in other machine learning contexts, such as predicting a defendant's future risks or misconduct, might make unfair judgments about which city areas are considered safe or unsafe. Historical bias can lead to situations that adversely affect communities that have been historically discriminated against. Therefore, it is crucial to ensure that the presence of robots in public spaces does not perpetuate or exacerbate stereotypes that society seeks to eliminate.}

\subsubsection{Aggregation Bias}\label{sec:aggregation} occurs when \change{incorrect conclusions are drawn for one group of people based on observations made from other groups}~\cite{suresh2019framework, mehrabi2019survey}. For example, recent work has demonstrated evidence of bias in the pedestrian recognition system of self-driving cars~\cite{ramanagopal2018failing}. \change{While deep learning approaches for object detection have facilitated significant advancements, the lack of diverse data from certain populations, such as people of color, can lead to misidentifications. This is primarily due to the concentration of technological development and data collection in geographic areas with predominantly lighter-skinned communities, which affects the accuracy of recognition for all the other diverse groups.} As a solution to this type of bias, in \cite{wilson2019predictive}, the authors present results from experiments \change{involving} eight AI models used in state-of-the-art object detection systems. \change{They also introduce the concept of “predictive inequity” for detecting pedestrians with different skin tones}.

\subsubsection{Algorithmic Bias}\label{sec:algorithmic} Algorithms play an essential role in the functioning of robotic systems, and algorithms can amplify data biases in the data. The study of algorithmic bias concerns algorithms that reflect systematic and unfair discrimination. Some researchers~\cite{rai2020explainable} argue that algorithms are neutral and unbiased as algorithms are treated as black boxes. However, algorithms can acquire human biases~\cite{corso2020survey}. Ethical research advocates for algorithms' transparency \change{to solve the bias problem}.

\subsubsection{Evaluation Bias}\label{sec:evaluation} arises during the evaluation of algorithms. This assessment is typically carried out using benchmarks that disproportionately and incorrectly measure the performance of algorithms~\cite{hurtado2021learning}. For instance, bias can be acceptable if its use is \change{appropriately} justified or unacceptable if it replicates, promotes, or amplifies social discrimination. In \cite{hurtado2021learning}, the authors present an example of evaluation bias using an algorithm that implicitly incorporates social dynamics through a learned model to guide a mobile robot to a destination. The algorithm subdivides the personal space of people into a human-robot interaction zone and a private zone where the robot cannot navigate~\cite{patompak2019learning}. As the algorithm's goal is to choose an optimal navigation strategy, the bias leads to robots having a greater distance from women than men. In this paper, the authors argue that there are no reasons to justify this gender bias that affects the robot's interaction with women.

\subsubsection{Developers Bias}\label{sec:developer} biases arise not only due to data but also from the developers or creators. This type of bias can be categorized as implicit and explicit bias. Implicit bias refers to software or robot engineering choices (e.g., feature selection or design choices regarding the appearance of a robot) that are not explicitly motivated or discussed but are induced by developers acting on cognitive biases influenced by their individual beliefs, actions, cultural and societal context. These biases indirectly affect how models are adapted~\cite{johnson2020algorithmic}. Explicit bias is sometimes consciously put into models by their developers, aiming to reduce certain types of inequalities or under-representation in data samples. Recent research~\cite{spatola2020implicit,gupta2018robot} has also exposed that explicit bias often comes from a lack of diversity in research groups in robotics. Consequently, a small homogeneous group decides the type of data, societal and demographic characteristics, and populations to be analyzed.

\subsubsection{Societal Bias}\label{sec:societal}  \change{occurs} when judgments or actions of people influence the behavior directly from users. When robots are deployed in the real world, they can indirectly learn biases that directly come from society~\cite{friedman1996bias}. They also reflect the personal biases of individuals who interact with the robot. This is particularly important because robots are not intended to replicate unethical aspects of society. Their participation in society is to be machines that help people to perform tasks or make their lives easier. To fully accomplish this task, social biases must not influence their behavior.

\subsection{Types of Discrimination}
\label{sec:discrimination}

\change{All different types of bias can lead to discrimination.} Discrimination is not static; \change{its causes and manifestations vary in a social context. Its resilience is rooted in the self-perpetuating nature of power and the inherent or social biases carried out by individuals and groups of people. Therefore, discrimination denies opportunities to members of one group that may be readily available to other groups. \change{Conversely,} bias is the preference that individuals have for certain groups of people, objects, and social contexts \change{based on their interests. Consequently}, actions stemming from these preferences can be interpreted as instances of discrimination. Some of this bias can occur unintentionally, while personal interests, cultural backgrounds, and the broader concept of cultural relativity \change{may drive others.}}

In robot learning, discrimination results from bias, and its consequences are linked to discriminatory behaviors in robots. This makes the robot more efficient for a particular group of people than other groups. Since robots are expected to be in critical human scenarios~\cite{arkin2011moral}, developers need to consider the possible types of discrimination. Notably, there are many potential ways in which robots can be taught to do something immoral, unethical, or just plain wrong. If developers are aware of the potential discriminatory scenarios, they can account for how the robot should interact fairly and equitably. Although the possible discrimination scenarios in robot learning have not yet been studied in-depth, in autonomous mobile robots such as self-driving cars, there is recent work in this area~\cite{hin2016structural, liu2017three, gentzel2019classical, brandao2021discrimination, dietrich2021understanding}. Nevertheless, the types of discrimination can be extrapolated to other robot learning applications such as assistive and service robots. In the following, we describe the types of discrimination that may occur in robot learning due to biases. 

\subsubsection{Direct Discrimination} occurs when robots treat people unfairly considering their protected characteristics~\cite{hajian2012methodology}. In this case, discrimination arises from procedures that explicitly use sensitive attributes such as race, gender, socio-economic status, geographical area, and other factors, for tailored autonomous decision making. In self-driving cars, direct or intentional discrimination concerns seemingly justifiable, even permissible, forms of active discrimination that involve justified benefit to a reduced group of people~\cite{liu2017three}. However, \change{the algorithm should not use} protected characteristics to distinguish individuals. Research in this area has shown that equality of opportunity is a helpful concept to avoid discrimination in autonomous vehicles~\cite{brandao2021discrimination}.

\subsubsection{Indirect Discrimination} arises when robots do not explicitly use protected characteristics to guide their decision making, but instead rely on how they interact and classify different groups of individuals according to their protected characteristics~\cite{discrimination2019robots, hajian2012methodology}. In machine learning, indirect discrimination has caused, for example, certain groups of people to be excluded from a job or denied credit. In~\cite{zuiderveen2018discrimination} the authors present an example of how a machine learning system that predicts who is eligible for a bank loan and who is not can generate indirect discrimination. They showed that even if the training data do not contain information about protected characteristics, the machine learning system learns that people from certain postcodes were likely to default. Here it seems reasonable to assume that there is no discrimination. However, the postcode may correspond with racial origins. In this case, the decision is causing harm to people of specific racial origins. A similar scenario can occur in robot learning. For instance, in \cite{skewes2019social}, the authors present the example of a robot conducting job interviews. They aim to present a method to avoid possible bias called fair proxy communication. This form of communication seeks to eliminate implicit biases that produce indirect discrimination by recognizing them to inform new possible interventions to reduce the possible harm.

\subsubsection{Explainable Discrimination} \change{refers to justified discriminatory robot behavior and occurs, for example,} when developers use protected characteristics to influence robot decision making to achieve fair outcomes~\cite{discrimination2019robots,pasquale2017toward}. \change{One of the most crucial abilities a personal robot requires when engaging with humans is the ability to distinguish them to promote well-being and adapt its interaction according to the user's necessities and preferences~\cite{alvarez2012feature}.} However, these personal characteristics cannot cause segregation, and their use must be just. \change{When developers incorporate specific protected characteristics into the robot's decision making algorithms, they must ensure a robust justification for their application and assess the scenarios upon which their application is grounded. This is essential to achieve heightened advantages and guarantee non-discrimination for the users.  For instance, developing robots capable of recognizing individuals with disabilities will foster a sense of inclusiveness for these individuals while using them.} In \cite{setchi2020explainable}, the authors present an intriguing perspective on explainable robotics. They argue that \change{creating methods and algorithms for generating explanations is essential to ensure robots work safely and reliably with humans.} These explanations allow robots to operate in different scenarios and communicate their decisions to humans, further facilitating effective human-robot interaction.

\subsubsection{Unexplainable Discrimination} refers to unjustified and \change{unjust discriminatory robot behavior targeted at a group often composed of minorities.} More specifically, it \change{arises} when robot decision making is based on the user's protected characteristics such as gender, race, disabilities, religion, and other characteristics, \change{and the different treatment cannot be explained by objective and adequate reasons, therefore unexplained, unjustified and unjust.} This type of discrimination can be illegal in specific contexts~\cite{hajian2014generalization, pasquale2017toward, lichocki2011ethical}.

\section{Detection and Mitigation}
\label{sec:tech}
\begin{figure}
    \footnotesize
    \centering
    \includegraphics[width=0.47\textwidth]{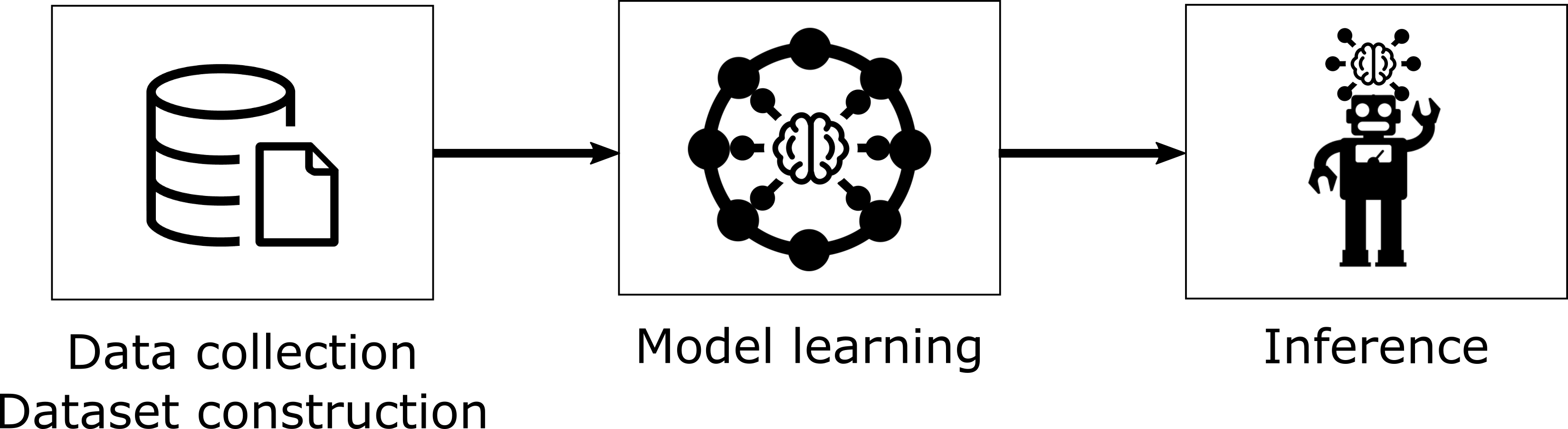}
    \caption{Typical simplified stages in the robot learning process. It consists of dataset construction or demonstration data collection, model learning, and inference. We use these stages to categorize the methods for fairness in robot learning.}
    \label{fig:robotlearning}
\end{figure}


Despite recognizing fairness as an essential value, fairness in robot learning is still in the early stage of research. We find fairness concepts only included in a few cases in the different learning stages through all learning techniques~\cite{brandao2019age, brandao2020fair, brandao2021discrimination, hurtado2021learning}. More considerable research advances exist in algorithmic fairness~\cite{dunkelau2019fairness, kleinberg2018algorithmic, beutel2019putting}. Such an example is algorithmic fairness in natural language processing, where discrimination is more explicit, and techniques for prevention and correction are more explored~\cite{jacobs2020meaning, chang2019bias}.

In the specific case of robotics, ethical scenarios are studied in autonomous driving, given the high interest in this topic and the general concern regarding safety~\cite{wilson2019predictive, gauerhof2020assuring}. Moreover, different works on this topic analyze the social implications, moral dilemmas, and the requirements that build the confidence needed to incorporate autonomous vehicles in real-world scenarios and safely operate around humans~\cite{reed2016responsibility, johnson2019artificial, arrieta2020explainable, goodman2017european, de2008atlas, vandemeulebroucke2020ethics, riek2014code}. Some other works in robotics concentrate on the physical characteristics and the assigned task that a robot should perform and provide the ethical implications~\cite{haring2018ffab, ogunyale2018does, addison2019robots}. Our approach in this paper is to collect the methods used to measure fairness in learning models and include fairness awareness in the context that can be applied in different stages of the robot learning process. Since learning algorithms constitute such an essential component in robot development, this work aims to provide an initial guide specifically for learning methodologies that contribute to widely benefiting humans across their demographics, characteristics, and diversity.

\begin{figure}
    \footnotesize
    \centering
    \includegraphics[width=0.37\textwidth]{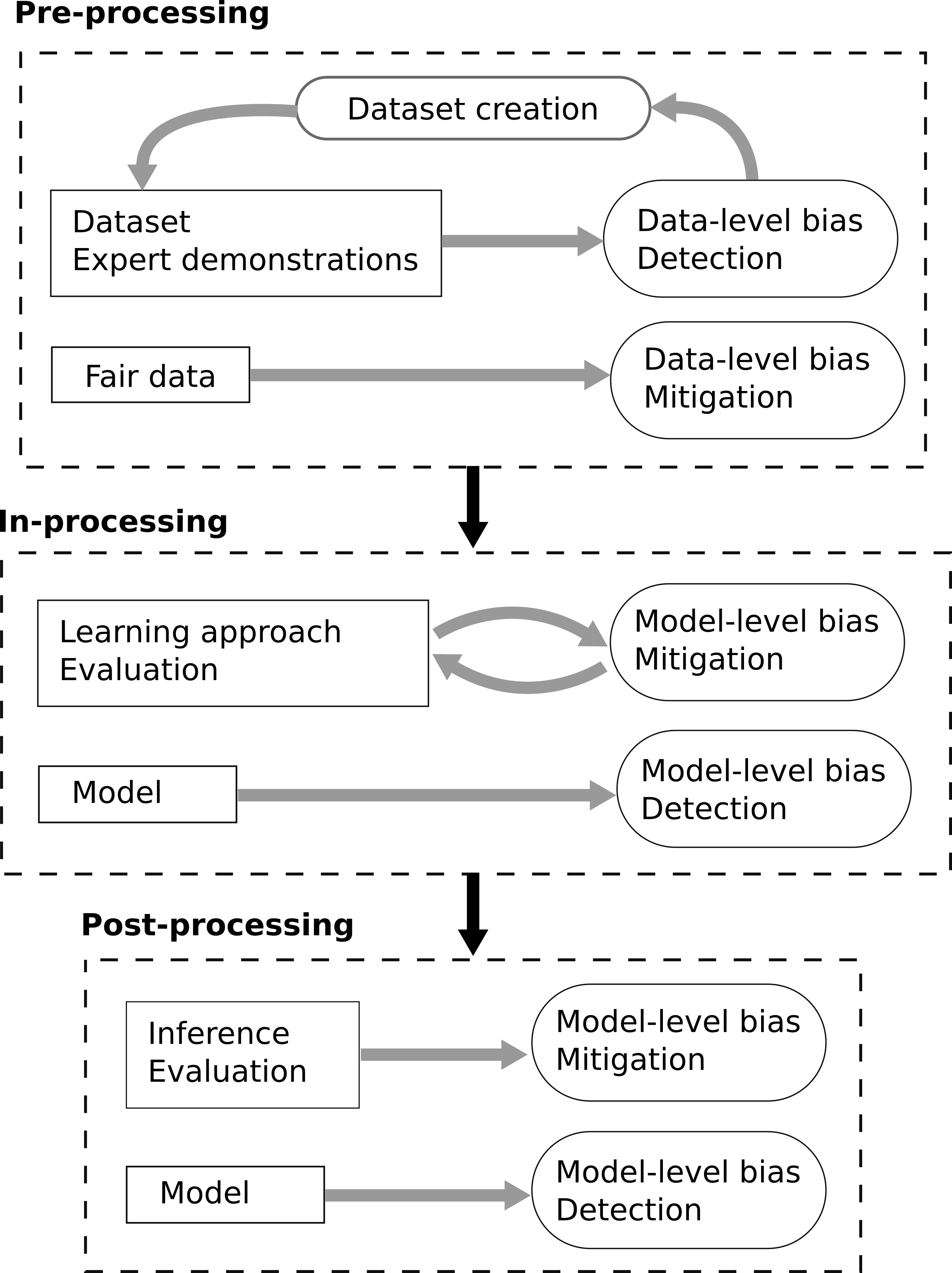}
    \caption{Block diagram illustrating the different levels where fairness can be included in the robot learning process.}
    \label{fig:robotlearning2}
\end{figure}

A typical and simplified robot learning process consists of three main stages: dataset construction or demonstration data collection, model training, and inference as shown in \figref{fig:robotlearning}. We previously identified bias at three levels in the robot learning process throughout the different learning techniques: data level, model level, and rectifying level. We use these stages prone to bias to categorize the explored methods for fair robot learning and review bias detection and mitigation strategies as shown in \figref{fig:robotlearning2}. We additionally survey different metrics proposed to measure model fairness and simulation and testing environments for fairness in robot learning.

\subsection{Fairness Metrics}

Measuring fairness in learning algorithms is a challenging task that remains an active area of research. Currently, there are no conventional fairness definitions or metrics for robot learning. Nevertheless, generating a single metric that satisfies fairness across social situations, cultures, and diverse individuals is unfeasible. Robots operating in real-world social scenarios must account for different perspectives to fulfill human safety and comfort standards. Thus far, multiple fairness metrics have been studied and proposed in the context of computer science and algorithmic fairness. Additionally, fairness metrics have been proposed in human-robot collaboration setups and as subjective questions to address fairness in the robot's performance.

Selecting the right fairness metric is another open challenge for fair robot learning. Not only can the fairness metric change across specific tasks, but it is also goal and context-specific. Besides the social-related difficulty, it has been shown that it is not feasible, from a technical point of view, to satisfy all fairness definitions simultaneously~\cite{kleinberg2016inherent}. An interdisciplinary approach supported by the definitions and explanations helps select the appropriate fairness metric and further interpret it. Accurate metric interpretation is key to localizing sources of bias and how to mitigate the effects. Additionally, fairness metrics should be diverse, dynamic, and easily adaptable to novel social scenarios~\cite{cerbone2021providing}. After evaluating different fairness metrics, Hinnefeld~\textit{et~al.}~\cite{hinnefeld2018evaluating} provide a set of warnings when selecting the appropriate approach. The authors found that in the cases where accuracy should be equal across groups of interest, most metrics are appropriate, and the interpretation is straightforward. Nevertheless, in the contrary case, such as in many robot scenarios, fairness metrics interpretation is challenging, and the intervention of human experts might be required.


\begin{table*}
\centering
\begin{tabular}{l|p{4.5cm}|p{5cm}|p{5.5cm}}
     \toprule
      & \multicolumn{1}{c|}{\textbf{Data Level}} & \multicolumn{2}{c}{\textbf{Model Level}} \\
     \cmidrule{2-4}
      & \textbf{Pre-processing} & \textbf{In-Processing} & \textbf{Implementation} \\
     \midrule
     \textbf{Detection} & \begin{itemize}[leftmargin=*]
         \item Dataset collection protocol~\cite{gebru2021datasheets}
         \item Data analysis, Label distribution~\cite{yang2020towards}
     \end{itemize}& \begin{itemize}[leftmargin=*]
         \item Fairness Metrics - Performance Quality~\cite{grgic2016case,feldman2015certifying,kusner2017counterfactual,hardt2016equality,buolamwini2018gender,chouldechova2017fair, li2023dark}

         \item Quality disparity \cite{du2020fairness,hurtado2021learning}
         \item Model Interpretation \cite{du2020fairness}
     \end{itemize} & \begin{itemize}[leftmargin=*]
         \item Fairness Metrics - Performance Quality~\cite{grgic2016case,feldman2015certifying,kusner2017counterfactual,hardt2016equality,buolamwini2018gender,chouldechova2017fair, li2023dark}
         \item Fairness Metrics - Collaborative Teams~\cite{chang2020defining,hoffman2019evaluating}
         \item Subjective Metrics for Fairness~\cite{buttner2023excuse,hitron2023implications}
         \item Test on real-world scenarios~\cite{kollmitz2015time}
         \item After deployment data collection~\cite{hurtado2021learning}
     \end{itemize} \\
     \midrule
     \textbf{Mitigation} & Data cleaning
     \begin{itemize}[leftmargin=*]
        \item Dataset cleaning \cite{du2018data}
        \item Fairness Through Unawareness \cite{grgic2016case}.
        \item Diversify data \cite{gupta2018robot}
        \item Data filtering \cite{dunkelau2019fairness,hagendorff2020ethical,agarwal2021does}
        \item Data augmentation \cite{calmon2017optimized, mclaughlin2015data}
        \item Unsupervised data analysis \cite{hurtado2021learning}
     \end{itemize} & \begin{itemize}[leftmargin=*]
         \item Regularization with social constraints \cite{kalweit2020deep,hamandi2019deepmotion}
         \item Regularization with fairness constraints \cite{kamishima2012fairness, kamishima2012fairness, iosifidis2018dealing,wilson2019predictive}
         \item Imitation learning~\cite{kalweit2020deep,hamandi2019deepmotion}
         \item Transfer learning~\cite{schumann2019transfer}
         \item Fairness-aware adversarial training \cite{zhang2018mitigating,madras2018learning}
     \end{itemize} & \begin{itemize}[leftmargin=*]
         \item Social adaptability~\cite{kollmitz2015time}
         \item Relearning~\cite{hurtado2021learning}
     \end{itemize}
     \\
     \bottomrule
\end{tabular}
\caption{Summary of different methods for bias detection and mitigation in robot learning. We classify the current methods into three different levels in the learning process: data level pre-processing, model level in-processing, and model level implementation, and the type of problem: detection and mitigation.}
\label{tab:my_label}
\end{table*}

\subsubsection{\change{Fairness Metrics Based on Performance Quality}}

\change{Various fairness metrics in ML that rely on performance quality such as the model's accuracy or error rate can be employed in robot learning. These metrics are particularly useful for assessing fairness with respect to the model's performance. In the following, we define these metrics and include examples in the context of robot learning.}

\begin{enumerate}[label=(\roman*)]
\item {Individual Fairness} is based on providing similar outcomes to similar individuals. Therefore, it measures the similarity distance between the individuals and the distance between the possibilities of obtaining outcomes assigned to them. Then, individual fairness estimates how distanced they are~\cite{zemel2013learning}. This can be measured by comparing input and output distributions with a statistical distance. However, previous work points out that individual fairness provides a problematic definition of fairness. The main concerns include the insufficiency of similar treatment and finding suitable similarity metrics~\cite{10.1145/3461702.3462621}.
\item{Group Fairness} is also known as statistical fairness and it is widely used as a fairness definition~\cite{awasthi2020beyond}. Different group fairness metrics are proposed based on sub-groups, usually chosen with sensitive attributes such as race or gender. These metrics use terms such as the rate of false positives (FPR) and false negatives (FNR). FPR is the probability of falsely obtaining a positive result and it is calculated as:
\begin{equation}
    FPR=\frac{FP}{FP+TN},
\end{equation}
where FP and TN are the False Positives and True Negatives, respectively. Similarly, FNR indicates the probability of being incorrectly assigned as negative and it is defined as:
\begin{equation}
    FNR=\frac{FN}{FP+TN}.
\end{equation}

By quantifying the incorrect results across different sub-groups, the following group fairness metrics indicate the presence and proportion of disbalance in the intended functioning of the algorithm.

\change{A related metric known as Miss Rate (MR) is commonly used to evaluate pedestrian detection performance. This metric is defined as }
\begin{equation}
    MR= 1-FPR.
\end{equation}
\change{It measures the percentage of pedestrians that go undetected. In a recent study on autonomous driving, the MR was utilized to identify a notable bias in current pedestrian detectors, negatively affecting children and individuals with darker skin tones \cite{li2023dark}.}
\end{enumerate}

\begin{itemize}
    \item \textit{Fairness Through Unawareness:} We can obtain a fair learned model by omitting explicit information from sensible attributes~\cite{grgic2016case}. With fairness through unawareness, we expect that the users in the subgroups $A=0$ to have an equal probability of being assigned the same model prediction distribution as the entire users, such that:
    \begin{equation}
    P(\hat{Y}|A=0) = P(\hat{Y}).
    \end{equation}
    
    Nevertheless, it has been shown that ignoring the sensitive attributes does not guarantee that the learned model is fair~\cite{hardt2016equality}. Even though this comparison is difficult to achieve, it is usually a good practice to prevent the inclusion of sensible attributes into the learning process unless a human expert has a valid reason to do so.

    \item \textit{Demographic Parity:} is a metric that assesses whether robot decisions have a similar effect across different groups of people. It is satisfied if the results of a learned model do not depend on protected attributes~\cite{feldman2015certifying,kusner2017counterfactual}. It is formally defined as:
    \begin{equation}
        P(\hat{Y}|A=0) =  P(\hat{Y}|A=1).
    \end{equation}
    
    With demographic parity, we aim that the subjects in groups $A=0$ and $A=1$ have an equal probability of being assigned to a correct positive model prediction $\hat{Y}$. Therefore, it promotes the True Positive Rate (TPR) to be the same for each subgroup such that:
    \begin{equation}
        TPR_{(A=0)}=TPR_{(A=1)}.
    \end{equation}
    
    In robotics, this can be correct functioning, completing the main goal, or adequately interacting with a user. For instance, a robot ideally interacts equally with users regardless of skin color. Therefore, demographic parity requires an equal proportion of positive robot performance throughout the groups of interest. However, achieving this absolute equal proportion in the real world is not feasible. Instead, an approach that aims to minimize the TPR gap is accepted, or a threshold $u$ is accepted such that:
    \begin{equation}
        \frac{P(\hat{Y}|A=0)}{P(\hat{Y}|A=1)} \geq u,
    \end{equation}
    where, as disparate treatment, it is usually accepted that $u=80\%$. 
    Demographic parity is helpful for reasoning about the inequities and monitoring the robot decision making process. However, the main risk in using demographic parity as a fairness definition is a generalization. For instance, if a user with a disability requires personalized interaction, the overall fairness evaluation with demographic parity will be affected \cite{hardt2016equality}. Therefore, satisfying democratic parity does not necessarily certify a fair robot model.
    
    \item \textit{Equalized Odds:} requires the positive outcome to be independent of the protected class $A$, conditional on $Y$ being an actual positive and it is defined as \cite{hardt2016equality}:
    \begin{equation}
    P(\hat{Y}|A=0,Y=y) =  P(\hat{Y}|A=1,Y=y), y \in \{0,1\},
    \end{equation}
    
    This metric compares both the true positive rates and false positive rates for different subgroups of interest such that:
    \begin{align}
        TPR_{(A=0)} &= TPR_{(A=1)}, \\
        FPR_{(A=0)} &= FPR_{(A=1)}.
    \end{align}
    
    This is a rigorous fairness metric since it aims to minimize the gap in both TPR and FPR. Therefore, the performance of the main task can be affected while increasing this fairness metric.
    
    \item \textit{Equal Opportunity:} promotes that each group should get a positive outcome at equal rates, \textit{assuming that there are reasons for individuals to belong to this subgroup.} Different from Equalized odds, this metric accepts that different groups can have different output distributions and it is defined as:
    \begin{equation}
    P(\hat{Y}=1|A=0,Y=1) =  P(\hat{Y}=1|A=1,Y=1).
    \end{equation}

    The main drawback when using this metric to evaluate the learned model is obtaining a high rate of false positives. In this case, the overall model accuracy decreases, affecting the performance of the main task. A clear example of this case is the gender and racial gap in facial recognition performance~\cite{buolamwini2018gender}. If we optimize the current model for equalized odds, the overall face recognition performance will decrease.
    
    \item \textit{Predictive Quality Parity:} also called predictive rate parity, allows for determining if the quality or precision rates are equivalent across different groups~\cite{chouldechova2017fair}. The quality is measured as the performance of the learned model by comparing predictions and groundtruth. We can use metrics such as accuracy, precision, recall, or F1 to obtain the quality rate, depending on the task. In robotics, for example, this metric is satisfied if the quality of the robot's actions is the same for different genders.
    
\end{itemize}

\subsubsection{\change{Fairness Metrics for Collaborative Teams}}

\change{One aspect of  HRI involves human-robot teams collaborating towards a shared objective. Within this context, more specific fairness metrics are proposed. Chang~\textit{et~al.} propose fairness as a measure dependent on the robot's level of effort and the fluency of its actions in human-robot teamwork\cite{chang2020defining}. Fluency refers to the smooth coordination of actions among teammates, resulting in a high degree of synchronization \cite{hoffman2019evaluating}. On the other hand, effort denotes the robot's deliberate attempts to accomplish its tasks. In this setup, \cite{chang2020defining} suggests evaluating fairness by considering workload, capabilities, and task types within a human-robot collaborative setting.}

\change{In terms of fairness, an optimal workload distribution is relevant given that excessively high or low workloads can adversely affect performance. The authors introduce the Equality of Workload metric denoted as $Ew$, which is computed as follows:}
\begin{equation}
Ew = \frac{\#CR - \#CH}{\#A},
\end{equation}
\change{where $\#CR$ is the number of subtasks completed by the robot, $\#CH$ is the number of subtasks completed by the human, and $\#A$ is the total number of actions or tasks. $Ew$ aims to equalize the distribution of subtasks among team members, regardless of the task type and their capabilities.}

\change{Similarly, it is crucial to allocate tasks based on team members' capabilities. The Equality of Capability metric $Ec$} can be used for this measure, which is defined as:
\begin{equation}
Ec = \frac{\#(CR \cap SR)}{\#SR} - \frac{\#(CH \cap SH)}{\#SH},
\end{equation}
\change{where $\#CR$ and $\#CH$ are the unique sets of subtasks completed by the robot and human respectively, $SR$ and $SH$ are the sets of strengths for the robot and human, and the intersection operation $\cap$ denotes the common elements between two sets. This metric aims to balance the distribution of subtasks that align with each team member's skills, such as task completion time, accuracy, throughput, and task difficulty levels.}

\change{Lastly, fairness can also be related to the equal access each team member has to specific types of tasks. This can be measured with the Equality of Task Type metric $Et$ and is defined as follows:}
\begin{align}
Ei &= \frac{\#(CR \cap Ti) - \#(CH \cap Ti)}{\#Ti} ;\,i = 1,2,...,n,\\
Et &= \alpha ^ T[E1,E2,...,En],
\end{align}
\change{where $Ei$ is the equality value of each task type,
$CR$ and $CH$ are the unique sets of subtasks completed by the robot and human respectively, $Ti$ represents a set of the same task type, and $\alpha$ denotes weights of importance for each task type with the weights summing to 1. This value attempts to equalize the distribution of subtasks across various task type categories among team members. This metric ensures that all team members have equitable opportunities and cost-sharing in the context of task assignments.}

\subsubsection{\change{Subjective Metrics for Fairness}}

\change{Fairness is a subjective concept influenced by individual, contextual, and cultural influences. Therefore, one approach to assess this subjective aspect of fairness is to conduct user studies using questionnaires. These surveys collect information regarding how participants perceive the robot's behavior, performance, and overall decision making concerning fairness \cite{buttner2023excuse}.}

\change{Questions that might be included in the surveys are:
\begin{itemize}
    \item “Overall, how much do you trust the robot?”
    \item “How was the overall experience?”
    \item “How would you define the role of the robotic agent?”~\cite{hitron2023implications}
\end{itemize}}

\change{A study in the field of service robots explores the extent to which efficient resource delivery by a service robot influences perceptions of unfairness and how this perceived unfairness affects human-robot interaction. The authors found that, even when a robot performs efficiently from a technical perspective, individuals may still perceive it as unfair, particularly when its actions lead to negative personal consequences. As a result, this perception of unfairness has a negative impact on the overall perception of the robot~\cite{buttner2023excuse}.}

\subsubsection{\change{Simulation and Evaluation Frameworks for Robot Learning}}
\change{Simulation and evaluation frameworks play a crucial role in developing robots capable of operating accurately and safely while following social norms. These frameworks offer a controlled and adaptable training and testing environment, enabling the assessment of factors such as dataset composition, task design, and algorithmic choices' impact on biases and fairness issues in robot learning models.}

\change{Although existing frameworks may not explicitly address bias and fairness scenarios, they hold great potential for future research in this domain. These frameworks create a controlled and adjustable environment that defines robot behavior, potentially promoting ethical considerations and reducing harmful biases. For example, consider a scenario where a robot is being trained to assist individuals in a healthcare setting. A well-designed simulation can replicate various healthcare contexts, including diverse patient demographics, medical conditions, and cultural sensitivities. By incorporating a range of patient profiles and situations, the simulation allows the robot to learn and adapt its responses to align with fairness and equity principles.}

\change{Simulation frameworks also offer a distinct advantage in bias reduction. In the context of autonomous vehicles, simulations can replicate challenging road conditions, diverse traffic scenarios, and pedestrian behaviors. This enables the intentional introduction of bias into the data to assess the vehicle's response to biased inputs. Exposing the model to such simulation scenarios makes it possible to fine-tune models to identify and mitigate biases in real-world situations. This iterative process of identifying, simulating, and addressing biases is essential for achieving fairness in robot learning.}

\change{A primary challenge in robot learning is the high cost and potential risks associated with real-world experiments. To address these challenges, simulation environments provide valuable solutions. These virtual setups allow robots to interact with diverse scenarios, offering a safe and controlled setting for learning and evaluation. Notable works in this domain include OpenAI's Gym \cite{brockman2016openai} and NVIDIA's Isaac Sim \cite{liang2018gpu}, which provide environments for training and evaluating robot learning algorithms. An example of simulation for robot learning that combines social notions is proposed by Carrasco~\textit{et~al.}~\cite{carrasco2023towards}. They introduce a method for assessing spatial and temporal robustness by simulating noise in the perception system within a comprehensive evaluation framework that considers accuracy, diversity, and adherence to traffic rules. Similar approaches can be extended to address fairness in robot learning.}

\change{Additionally, the work of Zhang~\textit{et~al.}~\cite{zhang2023learning} presents an approach called Reinforcing Traffic Rules (RTR), which combines imitation learning (IL) and reinforcement learning (RL) to train traffic agents in realistic traffic simulations. RTR aims to strike a balance between human-like driving behaviors and adherence to traffic rules, demonstrating improved performance in both standard and unconventional scenarios compared to traditional IL and RL methods. Another contribution to safety in autonomous driving is Adv3D, a closed-loop sensor simulation framework for evaluating self-driving vehicle autonomy performance. This framework emphasizes the impact of realistic vehicle shape variations on perception, prediction, and motion planning, highlighting the effectiveness of closed-loop testing in identifying and assessing scene appearance variations affecting autonomy performance. More generally, Yang~\textit{et~al.}~\cite{yang2023learning} discuss the potential of generative models in creating a universal simulator (UniSim) for real-world interactions. This approach employs diverse datasets to emulate human and agent interactions, enabling zero-shot real-world transfer for various applications, including vision-language planners and video captioning models. In the context of robot learning and ethical evaluation, UniSim highlights the possibility of using generative models to create a universal simulator that can train robots to interact with the real world based on diverse datasets, facilitating ethical evaluation and transferability of learned skills.}

\subsection{Data Bias Detection}
\label{sec:datalevel}

Discrimination in data can arise from the content of the data that reflects human biases and due to incomplete or unrepresentative information. In this direction, a previous work provides datasheets with a helpful protocol to follow when creating a dataset at the collection time~\cite{gebru2021datasheets}. On the other hand, in data bias detection, the goal is to analyze the data to find possible existing patterns of structural discrimination and the disproportion rate of samples of subgroups of people. For instance, as documented in~\cite{yang2020towards}, ImageNet~\cite{deng2009imagenet}, one of the most prominent vision datasets, contains demographic bias. To detect this bias, the authors performed a demographic annotation and analysis. Statistical analysis such as comparing label distribution regarding sensible attributes in the training data is a helpful tool for understanding discrimination in data. 

\subsection{Data Bias Mitigation - Preprocessing}

The goal of data level bias mitigation is to debias and improve the quality of the training dataset. Therefore, these bias mitigation approaches aim to refine the sample distributions of sensible attributes, filter samples that enforce bias, or conduct data modification or augmentation to diminish discrimination due to the training data~\cite{kamiran2012data}. The main idea is to process the data to obtain a clean or restored version of the dataset for later use in the robot learning process~\cite{du2018data}. A naive approach \change{omits sensitive attributes, so the model never uses them as input features and deletes the biased output labels}. This approach is called Fairness Through Unawareness and assumes that by ignoring the sensible attributes, we can obtain a fair learned model~\cite{grgic2016case}. The main limitation is that many highly correlated features \change{that are representative of sensitive attributes might exist.}

Unwanted outputs and discrimination towards diverse people in robot learning models are potentially provoked by the difference in label distribution in the training data. This unbalanced data is likely to \change{affect the most unrepresented individuals significantly.} To alleviate this representation bias, we aim to generate a balanced training dataset to increase the prediction quality of underrepresented groups by enforcing the dataset diversity. Therefore, datasets can be validated to understand if they contain potentially harmful biases or patterns that could lead to undesirable behavior. Different from the fairness context, \cite{gupta2018robot} presents a framework to reduce dataset bias for grasping tasks. Their results show that current learning approaches can amplify biases in data, preventing the model from learning truly generalizable models. The authors highlight the importance of diversifying the data for robot learning. A similar approach can be defined to reduce bias towards fairness. For complex and risky tasks such as autonomous driving, the diversity of the data is a crucial element in learning generalizable models that can drive in all different environments around diverse individuals. Nevertheless, it is important to note that balanced datasets alone do not guarantee a solution for representation bias. Recent works demonstrate that even training a model with balanced data can lead to learned models that still recover in their intermediate representation attributes such as race and gender~\cite{wang2019balanced, elazar2018adversarial}.

Data filtering consists of removing the bias from the training data so that the model does not have to account for discrimination. Therefore, only fair examples are shown during training instead of using the available data, resulting in a fair model. The fairness-motivated selection and sorting of data for robot learning remain unstudied. Nevertheless, the main idea of data filtering in machine learning and artificial intelligence is that the downstream task can use the “cleaned” data representation and produce results that preserve fair qualities. Fairness quality can be measured with metrics such as demographic parity and individual fairness~\cite{dunkelau2019fairness}. In robotics, such a downstream task can be planning a navigation path for rescue or selecting a person to interact with. Therefore, these data filtering approaches in robotics aim to remove information from the dataset that might result in unfair decisions while trying to modify the original dataset as little as possible. As a result, the primary robot task performance is not affected during the learning process.

Consequently, for a biased dataset $D$, a repaired and cleaned dataset $\Hat{D}$ is constructed which ignores the original bias-related information but is still as similar as possible to the original data $D$. In this direction, \cite{hagendorff2020ethical} presents a selection process for training data that improves the data quality in terms of ethical assessments of behavior and influences the training of the model. In their work, they present different use cases; one such case is self-driving cars. In terms of discrimination, the author states that biases are acceptable if they are critical for the legitimate solution of a given task. So the author proposes to promote biases in datasets used to train machine learning models that lead to a preferability of desirable features from an ethical point of view. Similarly, \cite{agarwal2021does} presents a solution to obtain a fairer subset of the samples. The authors find that cleaning the data helps make model predictions fair by balancing the true positive rate for the protected class across groups while maintaining the model's overall performance. We can expect similar results in robotics. Nevertheless, for robot learning, data filtering is particularly challenging given that not only the training data is filtered but also \change{newly incoming data}. Deploying robots in real-world environments implies that the robot encounters novel situations that can present biased data not considered to filter. Additionally, removing prejudices from the dataset or information that disproportionately favors or disfavors specific groups of people is a very challenging task, given that these aspects are difficult to identify.

Data augmentation refers to the process of creating new data samples using the information available in the training dataset. This process can increase the robustness of a model and prevent the model from overfitting. In this case, it is possible to replace fairness-sensitive features with alternative values that generate synthetic data to make the dataset more balanced where features and labels are edited to ensure group fairness~\cite{calmon2017optimized}. In this direction, \cite{mclaughlin2015data} proposes \change{using data augmentation} to reduce the negative effects of dataset bias for person re-identification. Nevertheless, research in this area still needs exploration.

The previous solutions for data level bias mitigation present an additional challenge given that they rely on demographic label annotation. In robotics, many datasets do not contain labels for sensible attributes. Moreover, obtaining such annotations in many robotic tasks is highly expensive and sometimes impossible. In this case, unsupervised methods such as clustering can be used~\cite{hurtado2021learning}.

\subsection{Model Bias Detection}
\label{sec:modellevel}

It is possible to detect bias in the model by measuring quality disparity. Biased models produce an imbalance in quality performance when comparing groups or individuals. The quality disparity is generally measured with a two-step method in the model learning process. First, the data is segmented into subgroups according to sensitive attributes. Second, we calculate and compare the accuracy for each subset. \change{Unlike} quality disparity, model interpretation is also proposed as a discrimination debugging tool that analyzes the learned representations~\cite{du2020fairness}. With model interpretation, their work aims to identify if a protected feature is encoded by the intermediate representation of a deep learning model and how the feature contributes to the model prediction. The reasons for discrimination are represented as class activation maps later used to investigate the models' regions of interest when making decisions. The activation maps visualization is then used as a tool for interpreting the model. For instance, in their work, they present an example in face recognition where the model focuses on the eye region for lighter skin groups while focusing on the nose region and chin region for darker skin groups. Nevertheless, model bias detection in robot learning is still unexplored.

More specifically in robot learning, \cite{hurtado2021learning} proposes to create a dataset with the robot experiences after deployment. Then this dataset can use clustering to analyze if the experiences are grouped and correlated to protected attributes. This is similar to the quality disparity but is achieved in an unsupervised manner. Additionally, \cite{brandao2019age} provides an analysis of pedestrian detection categorized by age and gender. The work explores the performance disparities of the first 24 top-performing methods for pedestrian detection. As a result, the author finds higher errors in children's detection. Similarly, the different fairness metrics and \change{the novel fairness concepts} can be used to detect bias in the learned model.

\subsection{Model Bias Mitigation - In-processing}

\begin{figure}
    \footnotesize
    \centering
    \includegraphics[width=0.47\textwidth]{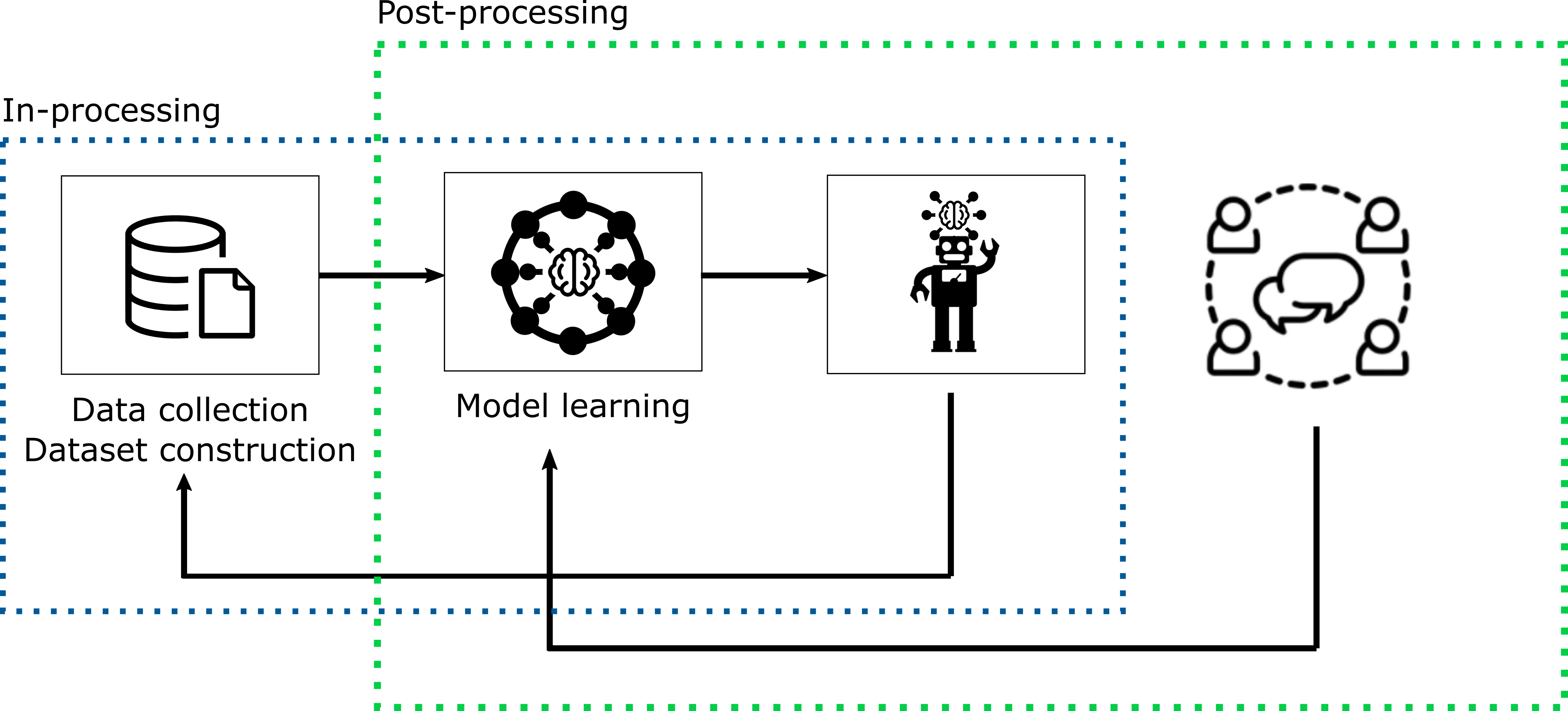}
    \caption{Learning a fair model for robots can include strategies categorized as in-processing and post-processing. First, in-processing strategies consist of adding fairness notions during the training stage. After obtaining a robot model, post-processing debiasing strategies aim to find and correct bias.}
    \label{fig:modelbias}
\end{figure}

The main objective of any learning process is to optimize the model to output a prediction as accurately as possible. In cases where a robot interacts with people or operates in social environments, we also require that robots perform safe actions around humans. Model Bias Mitigation approaches reformulate the robot learning problem and incorporate a fairness notion into the optimization process. Therefore, \change{the model is optimized for performance accuracy and fairness in model bias mitigation.} 

An evident solution is to change the loss function or to incorporate an auxiliary regularization term to the overall objective function during training, explicitly or implicitly enforcing constraints for specific fairness metrics~\cite{kamishima2012fairness, iosifidis2018dealing, wilson2019predictive}. For instance, \cite{wilson2019predictive} presents an approach that reweights the terms in the loss function to alleviate the skin color disparity in pedestrian detection. Therefore, their total loss function includes the weighting terms $\alpha_{LS}, \alpha_{DS}$ that apply to instances corresponding to light skin and dark skin, respectively. On the other hand, the goal of the regularization approach is to prevent the model from learning unwanted biases by adding a bias-related penalty term in the loss function~\cite{tartaglione2021end}. Such an example is the fair representation learning method presented by \cite{quadrianto2019discovering}. The authors include fairness in two components. First, it enforces learning representations that suppress protected attributes while performing the main task. Second, their method adds a regularization term to ensure the debiased representation lies in the same space as the original input. The fairness-related regularization can happen implicitly or explicitly. Implicit regularization adds constraints directly to disentangle the association between model prediction and fairness-sensitive attributes. It enforces models to pay more attention to correct features relevant to the prediction task rather than capture unwanted correlations between the prediction task and protected attributes~\cite{ross2017right, liu2019incorporating}. Differently, Explicit Regularization adds explicit constraints through the loss function to minimize the performance difference between different groups~\cite{agarwal2018reductions, kamishima2012fairness}.

Another possible approach consists of providing the robot with social skills. Inverse reinforcement learning is a technique used to train a robot model by learning a policy directly from human demonstrations to generate actions similar to human-like behavior. To include the social context in the learning process, these models aim to clone the behavior of humans. Subsequently, robots are equipped with these models for socially compliant actions~\cite{kalweit2020deep, hamandi2019deepmotion}. This approach has the potential for fairness-aware robot learning if we can guarantee that the demonstrations are bias-free. Nevertheless, this \change{complex task} will require the robot to learn only from a controlled and curated environment.

\change{Transfer learning is a different learning approach promising to diminish bias in the model.} This approach aims to pretrain a robot model using a source domain dataset. This source dataset is rich with data from a group of people underrepresented in the primary dataset. Then, the pretrained model is transferred to the primary dataset, and the learning process continues as fine-tuning. Transfer learning can improve the general performance accuracy and the performance accuracy for the underrepresented~\cite{schumann2019transfer}.

Process-based explainability is another promising approach for fairness-aware robot learning. Explainability is a challenging task in learning systems that consists of providing information that can clarify the input-output relationships. It is an essential step towards transparency in robotics. In general, for robot learning, we want explainability to include perception, decision, and action data to inform the reasoning behind the robot's behavior. Ideally, this allows for reconstructing an event or accident and highlighting changes to prevent unwanted robot actions. 
\change{To ensure} fairness, the explanation can trace the design and the implementation elements to ensure fairness~\cite{omeiza2021explanations}.

In model level training, adversarial approaches are also a possible solution to learning the main task without relying on information about sensitive attributes~\cite{zhang2018mitigating}. \cite{madras2018learning} presents an approach for deep learning models where both predictor and classifier are learned together. In this case, the predictor is trained to learn the main prediction task, while the classifier penalizes the learned representation if the considered sensible attribute is predictable. However, more research is needed to apply this approach in the context of robot learning.

\subsection{Model Bias Mitigation - Implementation}

Adaptability is an important attribute for robots to operate in the world. This allows the robots to appropriately function across diverse characteristics of specific users in the social environment where it is deployed. Therefore, in robot learning, post-processing approaches are critical. In this case, we recognize that the learned model may be unfair even if we include bias mitigation approaches in the data and model learning. Therefore, post-processing approaches update or transform the model to improve it in terms of fairness. This process is also called calibration or relearning~\cite{hurtado2021learning}, and its goal is to amend bias during inference time. Therefore we can enforce prediction distribution to approach either the training distribution or a specific fairness metric. These methods allow for diverse fairness metrics and prove effective in reducing discrimination.

The learning-relearning approach~\cite{hurtado2021learning} aims to include social and fairness notions into the different learning stages for robot navigation. Therefore, in this framework, socially aware robot navigation requires planning motions according to social norms and socially compliant behavior. \change{In this context,} a fairness point of view refers to promoting social good and avoiding prejudice through awareness and respect for human differences and necessities. Learning incorporates social context into learning navigation strategies so that robots can navigate in a socially compliant manner. Relearning diminishes any bias in the planned paths with the learned navigation model. The learning component can be obtained through inverse reinforcement learning, which allows for capturing the navigation behavior of pedestrians. Then, for relearning, a reward or punishment system in an off-policy reinforcement learning algorithm optimizes an augmented reward that encodes the detection of unfair behavior.

Additionally, social adaptability is also a potential factor related to fairness. In this case, \change{the goal is for robots to operate with human-like behavior.} One such approach for robot navigation is proposed by~\cite{kollmitz2015time}. It uses predicted human trajectories and a social cost function to plan collision-free paths that consider human comfort.  Additionally, their work uses time-dependent kinodynamic path planning to \change{consider the pedestrians' motion} around. In their work, the robot changes its navigation behavior to adapt to environmental changes. This principle can be projected into fairness motivation. Their work also highlights the relevance of testing in the real world to learn adaptability in social skills. Similarly, for fairness adaptability, real-world experiments are crucial before deployment.

\section{Legal and Fairness Considerations}
\label{sec:legalandfair}

In social science and philosophy, ethics is \change{regarded as an intricate system of moral principles and social rules that facilitate the assessment of an individual's actions as either good or bad~\cite{gert2002definition,abney2012robotics}. However, ethical explanations regarding human behavior have evolved throughout history. Consequently, what today’s society considers ethical vastly differs from the conception that people in the past held about it. The field of practical ethics is one of the outcomes of this evolution. This branch of ethics is typically applied to various fields of knowledge, whether technical or social. An example of this is the emerging field of roboethics~\cite{muller2020ethics}, which is tied to questions such as how to ensure responsible robot systems, what potential risks they entail, and how to mitigate them, as well as the social and economic implications robots might bring, among other considerations.}

Recent works have proposed fairness and legal considerations to achieve ethical robots. In this section, we present an overview of the ongoing research being carried out in this area. We first present a \change{comprehensive analysis of current advancements in the emerging field of roboethics, and subsequently, we narrow our focus to two aspects of roboethics:} fairness in robot decision making and legal considerations in robotics.

\subsection{Roboethics}

The interaction between humans and robots in social and personal environments has led to extensive studies on the ethical implications \change{surrounding} the development and use of robots. \change{Integrating ethical considerations during the development of robots intended for human interaction is imperative, as it helps to mitigate potential future risks, including instances of discrimination and segregation, all these concerns have given rise to the emergent field of Roboethics.} Research in roboethics \change{encompasses inquiries into achieving safety and preventing errors, as robots can be implicated in potentially fatal accidents. This area also delves into the legal and ethical dimensions linked to issues of responsibility for resulting harm. Ethical considerations within robot development, commercialization, and use are examined alongside topics such as fair robot decision making, privacy, the social impact of robots~\cite{veruggio2005birth, veruggio2016roboethics}}, human-robot interaction, and moral robot behavior. \change{It is important to note that aspects such as robots' consciousness, free will, dignity, and emotions are beyond the current scope~\cite{siciliano2016springer, veruggio2010roboethics, michalski2018sue,lichocki2011ethical}. Furthermore, the domain encompasses numerous other inquiries, including matters such as how to computationally model human moral reasoning in robotic agents~\cite{buechner2018two}.}

\subsection{\change{From Philosophy of Ethics to Robotics}}

\change{Roboethics involves exploring and integrating ethical theories such as deontology, consequentialism, egalitarianism, utilitarianism, and virtue ethics into the process of robot development. Importantly, these theories also function as guidelines for robot developers, ensuring the adoption of ethical practices. In philosophy, \textit{Deontology} is a normative ethical theory that contends the morality of an action should be determined by whether the actions themselves are right or wrong based on a set of predetermined rules [92]. Specifically, deontological principles clearly define which actions are ethical and which are not. A fundamental characteristic of this theory is that it emphasizes the importance of actions or inactions, rather than their consequences. On the one hand, in robotics, the ethical and critical stance of robot developers concerning the implications of their development on the world is regarded as a deontological application. On the other hand, Deontology in robotics also pertains to the practice of outlining robots' behavior based on algorithms and ethical standards~\cite{lin2012robot}. This involves explicitly determining how and why the attributes of a robot's ethical behaviors are chosen.}

\change{In a study conducted by~\cite{kim2021robots}, an applied form of deontology is presented that extends beyond designing a robot's moral behavior. The authors explore how robots can serve as moral advisors for humans based on deontological approaches. They argue that robots can encourage humans to make honest decisions under the premise that acting ethically holds greater significance than the consequences of their actions. The study includes human-robot interaction experiments, where the robots' role is to motivate individuals to behave fairly and honestly by offering moral advice.}

\change{Conversely, \textit{Consequentialism} belongs to the realm of normative ethical theories grounded in the principle that the outcome of actions serves as the basis for assessing their rightness or wrongness rather than the actions themselves~\cite{mansouri2016towards}. The fundamental distinctions between deontology and consequentialism in robotics lie in their applications: the former is used to learn a model for a robot's ethical behavior, while the latter is employed to evaluate the consequences of their behavior. Therefore, in robotics, this theory focuses on evaluating the moral implications of a robot's actions based on the consequences they generate rather than the actions themselves or any inherent rule. Therefore, developers should assess the potential consequences of a robot's actions and select only those that lead to the best overall outcomes for various stakeholders.}

Similarly, \textit{Utilitarianism and Egalitarianism} in roboethics refers to the development of fair robot decision making. \textit{Utilitarianism} is connected to consequentialism. According to this moral theory, an action is right if it promotes the happiness or pleasure of the greatest number. In robotics,  utilitarianism offers a method for deciding whether an action is moral or not. In other words, Utilitarianism in robotics refers to predicting various actions that robotic systems will perform,  estimating the benefits of these actions, and choosing only those actions that maximize well-being. \change{Utilitarianism is particularly popular in robotics due to its capacity for quantitatively evaluating the welfare or potential harm that robots can cause~\cite{wallach2008moral}. A crucial aspect of utilitarianism in robotics lies in learning models for robots to make choices that lead to the greatest benefit or positive impact for the largest number of people. For instance, in healthcare environments, a robot must prioritize patients who are more harmed or sick.}

\change{\textit{Egalitarianism} is the view that all individuals or social groups should have equal opportunities. In robotics, egalitarianism is used to develop fair algorithms that diminish inequality or discrimination in robot decision making through the lens of equality~\cite{brandao2020fair, zemel2013learning}. However, this concept is often criticized for failing to account for factors that people cannot control and morally justify indirect inequality. Since not all people are equal and have equal conditions, treating them identically can result in some obtaining more benefits than others.}
Finally, in~\cite{cappuccio2021can}, the authors proposed an interesting new approach to developing ethical social robots called \textit{Virtuous Robotics}. These robots have the task of helping people reach a higher level of moral development through \change{approaches rooted in virtue ethics.} As a result, they justify the use of virtuous robotics \change{by} illustrating this ethical application in robotics. 

\subsection{\change{Responsible Robot Development}}

Another significant point is that roboethics aims to ensure responsible robot development and robot decisions \change{align} with a set of universally shared moral values. \change{These values include} respect \change{for} human dignity and human rights, equality, justice, equity, benefit, respect for cultural diversity, pluralism, non-discrimination, stigmatization, social responsibility, etc. Following this scope, the EURON Roboethics Atelier project~\cite{veruggio2006euron}, created by scientists from the European robotics community, aims to \change{formulate} guidelines for building responsible robots \change{using} two strategies. \change{Firstly, the project seeks to create a roadmap} designed to assess opportunities to build and use advanced robot technology. \change{Secondly, it aims to evaluate the ethical challenges in robot development, comprehend these ethical dilemmas, and advance transdisciplinary research to address them~\cite{sullins2011introduction}.}

Finally, in [125], the authors from different fields propose five standards for the development and use of robots to achieve responsible robotics; these standards are \change{as follows}:
\begin{itemize}
    \item Robots should not be designed as weapons, except for national security reasons.
    \item Robots should be built and operated in accordance with current legislation.
    \item Robots are products that should be designed to be safe and secure.
    \item Robots as manufactured artifacts should not be used to exploit vulnerable users.
    \item Humans are responsible for any robot actions.
\end{itemize}

\change{The principles above were formulated to ascertain the necessary measures for integrating robotics research into society without causing harm.}


\change{Quantitative and qualitative research in social sciences can provide invaluable insights during development. There are several methods for assessing social acceptance and human perception of technology. The Technology Acceptance Method (TAM)~\cite{wu2011mixed} and The Unified Theory of Acceptance and Use of Technology (UTAUT)~\cite{williams2015unified} are used to predict the actual or potential use of technology, depending on the behavioral intention to use a system. This prediction corresponds to the relationship between behavior and behavioral intention. Behavioral intention refers to positive or negative feelings about a target behavior. The Pre-Prototyping Approach is generally used to test user's perception of the usefulness of technology. Other notable models in this realm include the Responsible Technology Acceptance Model (RTAM)~\cite{toft2014responsible}. This method combines self-interested motivation to use technology and moral reasoning to accept technology. For example, those models that cause social and environmental benefits are \change{readily} socially accepted.} 

\change{Similarly, in \cite{weiss2009usus} the authors analyze several factors for assessing the social acceptability of human-robot collaboration. These factors encompass usability, social acceptance, user experience, and societal impact. Building upon these factors, the authors introduce a novel theoretical framework known as USUS, designed to gauge the social acceptability of robots. This framework integrates methodological approaches from disciplines such as HRI, HCI, psychology, sociology, \change{and other disciplines}. The framework aims to ensure a positive user experience for all users, whether individuals or groups, to enhance social acceptance. Finally, in~\cite{livanec2022s}, the authors propose a method to foster collaborative knowledge production and reverse communication from the public to the scientific community, namely Cognitive-Affective Mapping (CAMs). This method is used to anticipate the psychological acceptance of robotics and its place within the broader context of modeling human attitudes and behavior. Therefore, CAMS is a tool that can be employed during the early stages of research to assess how individuals or groups are likely to accept or perceive robotics technology.}

\subsection{Fairness Definitions in Robot Learning}

Much of the literature on fairness in robot learning is motivated by the concern that high-impact decisions made by robots may have negative consequences or unfair distribution of tasks in human-robot interaction. Consequently, most of this work is focused on determining the meaning of this concept and how it should influence robots' behavior and decisions. \change{Paralleling the concept of roboethics, fairness in robot learning emerges as a multifaceted notion open to various interpretations. In this section, we present some of the research conducted thus far in the realm of fairness in robotics.} In \cite{otting2017criteria}, the authors \change{delve into} factors \change{essential for achieving} social interaction such as fair decisions in robot behavioral development. \change{They posit} that fair decision making \change{processes foster trust and improved social functioning and contribute to enhanced team performance. In support of this, they present a comprehensive overview of current design attributes aimed at cultivating optimal human-robot interaction. They assert that fairness becomes evident when robot decisions align with the principle of organizational justice \cite{colquitt2012organizational,greenberg2013handbook}.}

\change{This principle is rooted in the framework of decision making processes, encompassing elements of decision representation and distribution. Organizational justice~\cite{cohen2001role} is characterized by four dimensions: distributive, procedural, interpersonal, and informational justice. Distributive justice~\cite{lamont2017distributive} revolves around the equitable distribution of outcomes or decisions among stakeholders. Procedural justice \cite{lind1988social,nagtegaal2021impact} concerns impartiality in decision making procedures, including factors such as consistency, absence of bias, accuracy, correctability, and ethical considerations. Interpersonal justice \cite{liang2020interpersonal}  pertains to how robots communicate, emphasizing that robot behavior is perceived as fair when it is respectful, polite, and dignified. Informational justice is connected to transparency in robot decisions.} The authors \change{further} explain that research in this area has included the organizational justice principle in the robot's decision using fair algorithms.

In fair robotics, \change{the concept of} fairness is also \change{intrinsically linked} to cooperative \change{endeavors involving} humans and robots. In \change{a study documented in} \cite{9223594}, the authors \change{performed experiments to delve into a human teammate’s perception of fairness in the context of a collaborative task shared between humans and robots. During this collaborative effort, specific tasks were strategically assigned to capitalize on the respective strengths of the robot and the human team member. Notably, the study highlights the crucial role of meticulous design in achieving fairness within human-robot teamwork. Further, fairness serves as the foundation of trust and team effectiveness. \change{The same study},} \change{investigates} the effects of \change{two factors:} fluency (absent vs. present) and effort (absent vs. present) in collaborative tasks. \change{The experiment's results} \change{revealed that effort and fluency play a significant role in achieving fairness, all while maintaining efficiency. The authors also concluded that participants’ perceptions of fairness in a collaborative task are influenced by the skill levels of team members and the nature of the task itself. Building on this, they introduced three dimensions of fairness in human-robot teamwork:} equality of workload, equality of capability, and equality of task type. \change{Similar findings were presented in~\cite{el2014fairness}, where the authors emphasized that \change{these machines need to handle a broader spectrum of tasks for fair interaction between humans and robots.} More recently, \cite{londono2022doing} introduced an innovative approach to foster social and fair behavior learning in human-robot collaboration, grounded in \change{recognizing} anti-social and unfair behavior. The authors argue that} to avoid dilemmas in specifying and modeling sociability and fairness in robotics, learning to avoid unfair behavior can be an optimal way to achieve fair human-robot interaction.

\subsection{Legal Considerations in Robot Learning}
\label{sec:legal}

The increasing real-world applications of robotics have given rise to ethical and political discussions, and policy efforts to regulate the trajectory of these technologies. There is significant public discussion about artificial intelligence ethics, and there are frequent discourses from regulatory bodies that the matter requires a new policy, \change{primarily} due to the autonomy of these systems. Governments around the world, especially the United States of America, the European Union, Russia, and China, consider the development of artificial intelligence and robotics as economic and policy priorities, mainly because these systems can result in unfair and discriminatory behavior that can reinforce social inequalities~\cite{RoboLaw214robotics, wachter2017transparent, winfield2018ethical, Ethialrobots2016design, koops2013ethical}. However, drafting policies on this subject is complex due to infrastructure taxation, goodwill statements, regulation by various actors and the law, and funding~\cite{muller2020ethics}. Towards this effort, \change{scholars,} governments, parliaments, associations, and industry have presented proposals to regulate robot development, commercialization, and use~\cite{leenes2017regulatory,nagenborg2008ethical,winfield2019ethical,chatila2019ieee,palmerini2014guidelines, langman2021roboethics}.

As a consequence, there are promising advances in the subject. Most guidelines analyze liability, warfare, safety, discrimination, fairness, and human-robot interaction. Notably, regulating the behavior of those working in the robotics industry is also one of the main features of these guidelines~\cite{nagenborg2008ethical}. In \cite{muller2020ethics}, the authors argue that those who work on ethics and policy in robotics tend to overestimate its impact and threats and underestimate how far current regulation can reach. There are normative frameworks and legislations on fairness and non-discrimination. 

\subsubsection{\change{Human Rights Framework}}

Most importantly, there is a human right to equality and non-discrimination which is codified, for example, in Art. 26 of the International Covenant on Civil and Political Rights (ICCPR)~\cite{tomuschat2008international} and in regional human rights treaties, such as Art. 14 of the European Convention on Human Rights (ECHR)~\cite{waldock1958european}. The term equality can imply formal and/or substantive equality. Formal equality means primarily procedural equality, e.g., equality before the law and equality of the law~\cite{moeckli2010equality, voeneky2022cambridge}. Substantive equality implies equality of opportunities and equality of results, some call this equality through the law~\cite{moeckli2010equality, voeneky2022cambridge}. The human right to equality and non-discrimination primarily obliges states to grant equality before the law and equality of the law, e.g., the law should not discriminate. However, states may also have positive obligations to protect individuals from discrimination by regulating private conduct, especially regarding employment, education, or private activity in a quasi-public sector~\cite{moeckli2010equality, al1995law, schabas2019international}. It is important to note that states may also take affirmative action (positive discrimination) to overcome structural discrimination and establish equality of opportunity or equality of results (e.g., by introducing quotas). Considering the case law of the Human Rights Committee and the European Court of Human Rights, such forms of (positive) discrimination do not constitute a violation of Art. 26 ICCPR or Art. 14 ECHR, as they can be objectively and reasonably justified, i.e. the legitimate aim is to establish de facto equality~\cite{al1995law, schabas2019international}. \change{Several international conventions} specify the obligations of states \change{concerning} discrimination based on protected characteristics, e.g., the International Convention on the Elimination of All Forms of Racial Discrimination (ICERD)~\cite{unies2006international} or the Convention on the Elimination of All Forms of Discrimination against Women (CEDAW)~\cite{assembly1979convention}.

\subsubsection{\change{Anti-Discrimination Legislation}}

States have adopted legislation to protect individuals from discrimination, for example, the Civil Rights Act of 1964 in the United States or the Council Directive 2000/78/EC of the European Union regarding employment~\cite{directive2000establishing}. These laws could limit the discriminatory use of robots in employment settings. With regard to anti-discrimination law in the United States, e.g., the Civil Rights Act of 1964, researchers argue~\cite{xiang2019legal} that there is a gap between the definitions of fairness in machine learning and the actual legal concepts on which they are supposed to be based. This gap is due to the misunderstanding of legal concepts in the field of machine learning. The authors point out that while discrimination in machine learning is related to a false outcome, US anti-discrimination law defines discrimination not through outcome but causality and intent, which is difficult to apply to non-human decisions. 

Regarding the employer’s decision to use the robot, discriminatory intent can be \change{challenging} to prove because the employer will often not know that a discriminatory effect exists~\cite{barocas2016big}. Disparate treatment (direct discrimination) will thus be difficult to prove in US anti-discrimination law. Anti-discrimination law also protects against indirect discrimination (disparate impact), i.e. a treatment that affects people differently without considering protected characteristics, which is particularly relevant in algorithmic and robot learning because of the unintended discriminatory effects [123], [124]. In anti-discrimination law in the United States and the European Union, a disparate impact (indirect discrimination) can be reasonably and objectively justified~\cite{xiang2019legal, voeneky2022cambridge}. In~\cite{voeneky2022cambridge}, the authors argue that the legal requirements to justify indirect discrimination imply that minimum methodological standards must be met in machine learning (or robot learning). However, anti-discrimination law currently covers only certain settings, primarily employment, and there could be regulatory gaps regarding the use of robots in other circumstances.
Existing legal provisions set standards regarding equal treatment and non-discrimination. These standards need further elaboration, especially regarding whether the current criteria for determining discrimination can reasonably be applied to robots and whether and how discrimination can be justified. 

\subsubsection{\change{Further Legislation}}

\change{Data protection law could also provide for comprehensive standards regarding robot learning. Data protection law regulates the collection and processing of personal data. The General Data Protection Regulation (GDPR) of the European Union is one of the most advanced legislations~\cite{regulation2016regulation}. As outlined above, data sets in robot learning are mostly anonymized, and the protections for processing personal or sensitive data \change{therefore} do not apply~\cite{schreurs2008cogitas}. Art. 22 GDPR sets requirements for automated decision making. The meaning of this provision is still disputed~\cite{voeneky2022cambridge,kerrigan2022artificial}, but it sets some standards and limits that could be relevant for robot learning.}
\change{Considering approaches for the regulation of AI might also be useful.} Several ethical guidelines regarding artificial intelligence have been adopted, for example, by the OECD~\cite{council2019recommendation} \change{and the UNESCO~\cite{unesco2021recommendation}}. They are also concerned with non-discrimination and establish principles \change{applicable to} fairness in robot learning, such as transparency, explainability, safety, or security. These principles can also be found in the draft proposal of the Artificial Intelligence Act of the European Union \change{issued by the European Commission in 2021} (e.g., Art. 13, 15, 52)~\cite{act2021proposal}. The Draft Act establishes in Art. 10 requirements regarding data and data governance for high-risk AI, e.g., Art. 10 (3) states: “Training, validation, and testing data sets shall be relevant, representative, free of errors and complete”.

\change{Another important aspect of robot regulation, besides non-discrimination, \change{is} civil liability rules, thus the accountability for damage} caused by a robot, which is addressed, for example, in the European Parliament's resolution in "Civil Law Rules on Robotics''~\cite{Parliament2017robots}. Importantly, transparency, as a mechanism to explain robot behavior and decision making clearly, is presented as a fundamental tool to achieve accountability~\cite{European2021robotics, winfield2017case}. This issue has already largely been discussed in the context of autonomous driving cars~\cite{stilgoe2018machine,bonnefon2020ethics,boeglin2015costs,brodsky2016autonomous}, and some conclusions might be drawn from this for robot accountability. \change{Thus, some standards regarding robot learning can be derived from existing normative frameworks and legislations.} 

\subsubsection{\change{Regulatory Challenges and Future Pathways}}

\change{One also needs to consider that there might be regulatory dilemmas in robotics.} In~\cite{leenes2017regulatory, van2014ethicist}, the authors analyze four regulatory dilemmas in robotics, namely how to keep up with technological advances, how to balance innovation and human rights protection, whether to conform to existing social norms or to create new ones and how to balance efficiency with techno-regulation. In each regulatory dilemma, the authors focus on liability, privacy, and autonomy, which often \change{require} regulatory attention. In this paper, the authors also discuss that not only technology but also the adverse effects of technology should be regulated. Similarly, in other works, researchers claim that one of the main challenges of technology regulation is keeping pace with technological advances. In law, this is known as the pacing problem~\cite{marchant2011growing}. \change{As illustrated before in the case of the US anti-discrimination legislation, it can be difficult for lawyers and legal scholars to predict or quickly identify gaps in current regulatory regimes due to a lack of technical expertise. Conversely, most technical experts and robot developers do not know the existing regulatory framework, and there are uncertainties regarding the legal requirements. Interdisciplinary work combining technical and legal considerations allows us to fill the legal terms with life and foster responsible innovation (see for example~\cite{finck2021reviving}). It helps to clarify the technical implementation of legal principles concerning robot learning. For example, developing an appropriate methodology for robotic learning is necessary. However, it also helps to evaluate whether adapting existing legal concepts is sufficient, such as the current legal definition of discrimination in US law, and/or whether adopting new legislation is necessary. This could be one pathway to overcome the regulatory challenges to improve robot regulation and ensure 
} that the robot's creation, commercialization and use benefit people and do not infringe on human values, especially the human right to equality and non-discrimination.

\section{Practices for Fair Robot Learning}
\label{sec:practices}

\change{It is essential to prioritize human safety, comfort, and non-discrimination to develop reliable robots capable of learning models to manipulate objects in collaborative configurations, navigate social environments, and engage with humans. This involves a comprehensive exploration of the social context, considerations, implications, and a thorough analysis of the legal framework and the integration of technical strategies. In addition to these principles, specific practices might be crucial in promoting fairness in robot learning. Having guidelines for these practices aims to take care of details such as team creation, evaluation, and action protocols in case of unfair outputs in the learned model. Consequently, we present a set of initial practices to promote fairness-aware robot learning, with examples contextualized within the fields of robot learning and human-robot interaction.}

\begin{itemize}
\item Human-centered design: \change{The development of robot learning solutions should consistently incorporate a human-centered perspective throughout all phases~\cite{european2020artificial, wu2011mixed, williams2015unified}. In autonomous driving, a human-centered design approach involves considering the comfort and safety of passengers and pedestrians while maintaining accuracy standards through different groups. For instance, the model should prioritize smooth and safe driving behaviors to minimize discomfort and reduce the risk of accidents. Additionally, the model should remain effective in recognizing all pedestrians.}

\item Interdisciplinary perspectives: \change{Promote collaboration among experts from different fields to bring diverse insights into robot learning development and evaluation~\cite{boykin2021opportunities, Trustrobots,ramanagopal2018failing}. For example, when creating a navigation model for a robot in a healthcare setting, collaboration between engineers, healthcare professionals, and ethicists can lead to more fair and effective solutions, ensuring that the robot effectively operates while behaving safely following social norms.
}
\item Explicit guidelines and policies for inclusion and non-discrimination: \change{Establish clear guidelines to promote diversity within robot development teams, leading to more comprehensive and inclusive solutions~\cite{zhang2023fairness,directive2000establishing, xiang2019legal,barocas2016big}. Ideally, interdisciplinary and intercultural research groups will include different social perspectives to create empathetic solutions. For instance, research groups working on social robots should actively seek members from diverse cultural backgrounds and disciplines. This ensures that various societal perspectives are considered when developing robots interacting with diverse communities.
}
\item Optimization techniques: \change{Shift the optimization focus from mere efficiency to responsible robotics. Unlike optimization for efficiency and efficacy, explore approaches including constraints for optimization towards responsible robotics using fair decisions making methods~\cite{hurtado2021learning,kalweit2020deep, hamandi2019deepmotion,kollmitz2015time}. For example, when developing a robot learning model, it should incorporate fairness criteria to avoid biased decisions in sensitive domains such as healthcare. Moreover, autonomous driving systems can be optimized to acknowledge ethical decisions in complex situations such as prioritizing pedestrian safety.
}
\item Early detection and correcting unwanted robot behavior: \change{Explore learning frameworks that identify and penalize unwanted robot behavior during training~\cite{hurtado2021learning, shevlane2023model}. Implement and research testing environments with innovative metrics to evaluate and correct undesirable outcomes. In robot learning for human-robot interaction, if a robot learns to use disrespectful language, the framework should recognize and penalize this behavior, encouraging the model to adopt more respectful communication patterns.
}
\item Online evaluation to supervise models for responsible robotics: \change{Aim for models with adaptable skills, ethical awareness, and continuous online evaluation. Incorporate constant evaluation that includes new possible scenarios~\cite{hurtado2021learning, suchan2019out}. Developers can ensure their models evolve to handle diverse and changing social contexts by continuously evaluating robot behavior, particularly in novel and unforeseen scenarios. An illustrative example could be a robot developed for elderly care, which continuously adapts its behavior based on feedback from both residents and caregivers.
} 
\item Informed consequences and impact of the developments: \change{Developers should proactively assess the potential social impact of their robotic developments' and reach. This implies extensive consideration of the effects that the deployment of a robot may have on various scenarios~\cite{brundage2018malicious,mitchell2019model}. For instance, when introducing a robot into a workplace, careful consideration should be given to its potential impact on workplace dynamics. Specifically, while learning object manipulation, understanding the potential consequences of a robot's actions such as dropping fragile objects, allows for informed decision making and risk mitigation.
}
\item Conscious development: \change{Given that robots can physically interact with humans, developers should constantly be aware of the potential harm~\cite{shevlane2023model,veruggio2006euron,sullins2011introduction}. In the context of robot learning for navigation, developers should prioritize safety features that prevent collisions with humans or other obstacles, reducing the risk of harm.} 
\end{itemize}

\section{Conclusion}
\label{sec:conclusion}

In this survey, we exposed potential adverse issues that can affect robot learning systems due to bias and unfairness in the learning algorithms. We demonstrated the extent to which such problems can affect our society through examples of real-world scenarios. By analyzing the challenges and defining the problem from three perspectives, we presented the technical, ethical, and legal advances toward addressing fairness in machine learning for robotics. From the social focus, we created a taxonomy for different biases and categorized the resulting discrimination that can occur. We reviewed the legal and fairness considerations from an ethical and regulatory perspective. Inspired by recent advances in fair machine learning, we presented \change{bias detection and mitigation techniques} in robot learning algorithms. We showed that the model bias to some extent exposes biases present in our society and robot learning algorithms should reduce these biases instead of amplifying them. \change{This survey aims} to combine perspectives from different disciplines and provide tools, guidelines, and techniques for fair robot learning that will eliminate disparity and promote fairness in the field.

\section*{Acknowledgements}

This work was funded by the Carl Zeiss Foundation under the ReScaLe project and the German Research Foundation (DFG) Emmy Noether Program grant number 468878300.

\ifCLASSOPTIONcaptionsoff
  \newpage
\fi


\bibliographystyle{IEEEtran}
\bibliography{references.bib}

\begin{IEEEbiography}[{\includegraphics[width=1in,height=1.25in,clip,keepaspectratio]{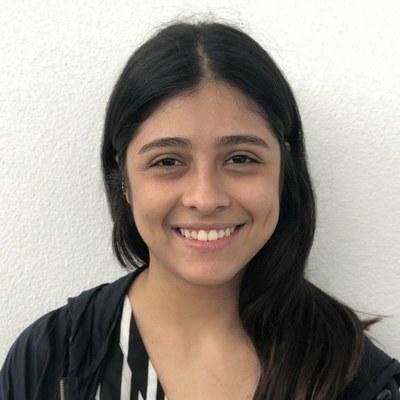}}]{Laura Londoño} is an M.A. Student in Interdisciplinary Ethics and an  Associate Researcher in the Robot Learning Lab at the University of Freiburg. She received her B.A. in Philosophy and Literature from the University of Caldas. Her research focuses on fair robotics, human-robot interaction, practical and interdisciplinary ethics, and robotphilosophy. 
\end{IEEEbiography}

\begin{IEEEbiography}[{\includegraphics[width=1in,height=1.25in,clip,keepaspectratio]{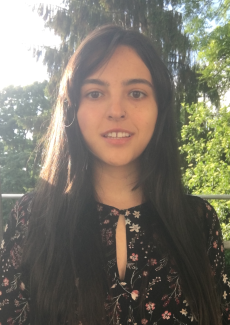}}]{Juana Valeria Hurtado} is a Ph.D. Student in the Robot Learning Lab at the University of Freiburg, and a scholarship holder in the Graduate School of Robotics. Her research focuses on learning-based methodologies for scene understanding and human-robot interaction with technical and social approaches. With her interest in robot learning and fairness, her aim is to equip robots with the ability to operate in social environments while meeting performance, safety, and fairness criteria.
\end{IEEEbiography}

\begin{IEEEbiography}[{\includegraphics[width=1in,height=1.25in,clip,keepaspectratio]{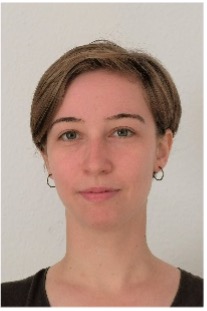}}]{Nora Hertz} is a Ph.D. candidate and research assistant at the Institute of Public Law Department 2 (Public International Law, Comparative Law, and Ethics of Law - Prof. Dr. Silja Vöneky) at the University of Freiburg. She studied law at Humboldt-University Berlin, Paris-Panthéon-Assas University, and King´s College London (LL.M.). Her research focuses on human rights and neurotechnologies, particularly the right to freedom of thought and the right to private life, as well as human rights-based technology regulation.
\end{IEEEbiography}

\begin{IEEEbiography}[{\includegraphics[width=1in,height=1.25in,clip,keepaspectratio]{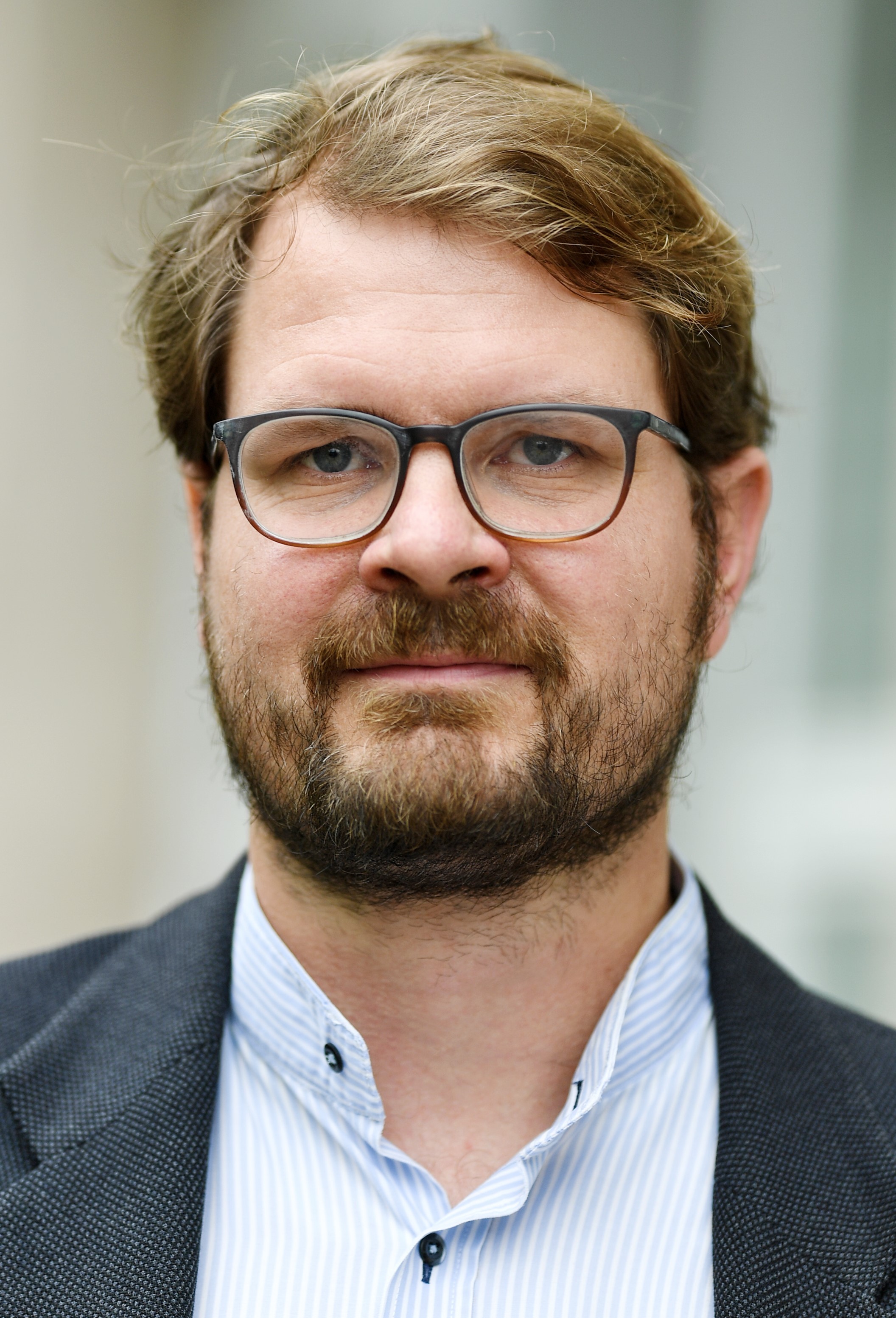}}]{Philipp Kellmeyer} is an Assistant Professor for Responsible AI and Digital Health at the University of Mannheim. He is also a neurologist and neuroscientist and heads the Human-Technology Interaction Lab at the Department of Neurosurgery, University of Freiburg Medical Center. He works on participatory approaches to responsible human-machine interactions in medicine and the application of digital health technologies. He also co-leads the Responsible AI research focus at the University of Freiburg.
\end{IEEEbiography}

\begin{IEEEbiography}[{\includegraphics[width=1in,height=1.25in,clip,keepaspectratio]{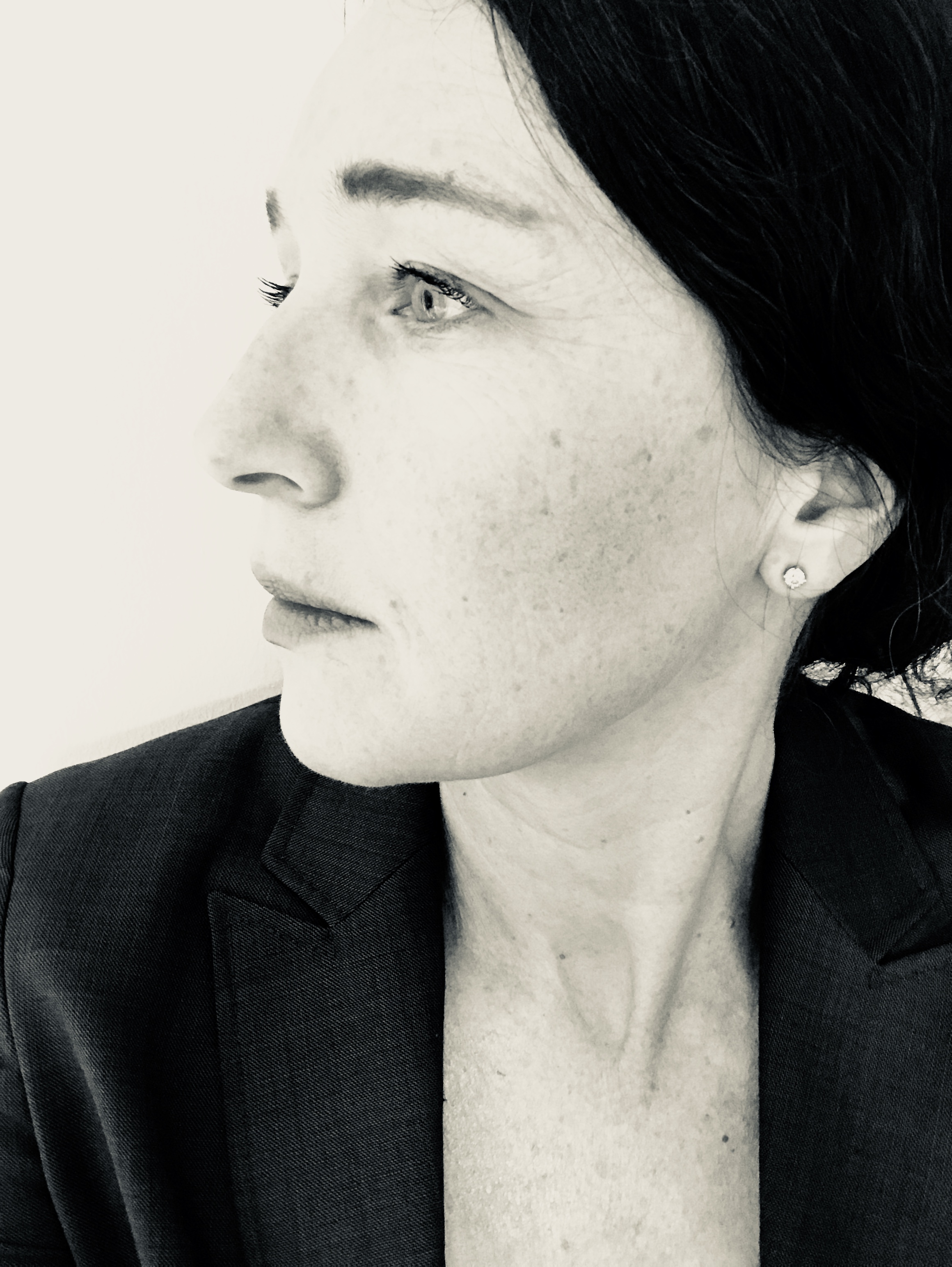}}]{Silja Voeneky} is a full Professor of Public International Law, Comparative Law, and Ethics of Law and former Vice Dean of the Law Faculty at the University of Freiburg. She was a Fellow at Harvard Law School (2015–2016) and a Fellow at the 2018–2021 FRIAS Saltus Research Group “Responsible AI - Emerging ethical, legal, philosophical and social aspects of the interaction between humans and autonomous intelligent systems”. Her areas of research include laws of war, environmental law, human rights law, and the interdependence of Ethics and Law, esp. questions on the Governance of disruptive science and technologies. She previously served as Director of an Independent Max Planck Research Group (Heidelberg 2005–2010), and a member of the German Ethics Council appointed by the Federal Government. Since 2001, Voeneky has been a legal advisor to the German Federal Foreign Office, the German Federal Ministry of Research, and the German Federal Ministry of Environment.
\end{IEEEbiography}

\begin{IEEEbiography}[{\includegraphics[width=1in,height=1.25in,clip,keepaspectratio]{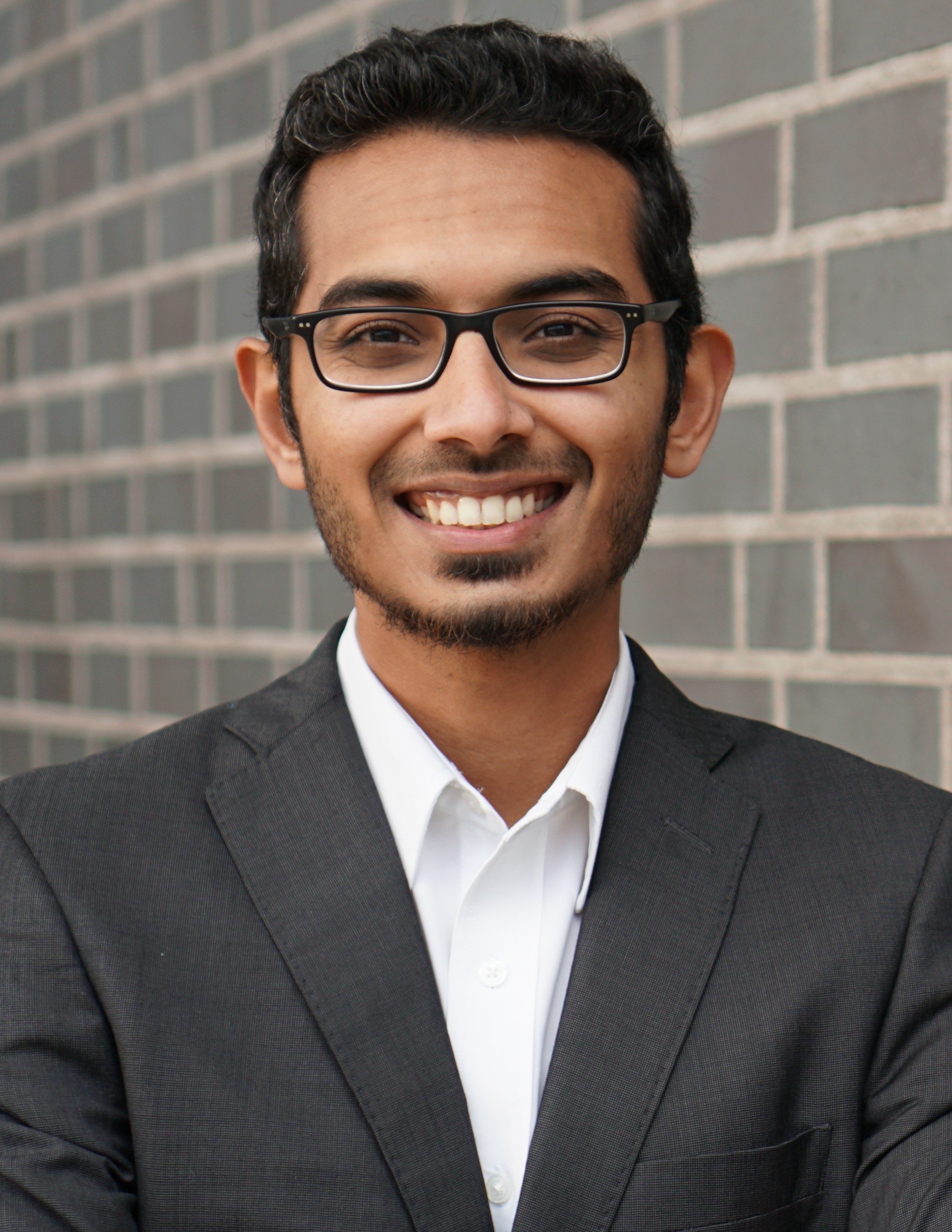}}]{Abhinav Valada}
is a full Professor and Director of the Robot Learning Lab at the University of Freiburg. He is a member of the Department of Computer Science, a principal investigator at the BrainLinks-BrainTools Center, and a founding faculty of the European Laboratory for Learning and Intelligent Systems (ELLIS) unit at Freiburg. He received his Ph.D.~in Computer Science from the University of Freiburg in 2019 and his M.S.~degree in Robotics from Carnegie Mellon University in 2013. His research lies at the intersection of robotics, machine learning, and computer vision with a focus on tackling fundamental robot perception, state estimation, and planning problems using learning approaches in order to enable robots to reliably operate in complex and diverse domains. Abhinav Valada is a Scholar of the ELLIS Society, a DFG Emmy Noether Fellow, and co-chair of the IEEE RAS TC on Robot Learning.
\end{IEEEbiography}


\vfill


\end{document}